\documentclass[10pt,journal,compsoc]{IEEEtran}
\usepackage{booktabs}
\usepackage{color}
\usepackage{xcolor}
\usepackage{times}
\usepackage{epsfig}
\usepackage{graphicx}
\usepackage{graphics}
\usepackage{amsmath}
\usepackage{amssymb}
\usepackage{multirow}
\usepackage{mathrsfs}
\usepackage{rotating}
\usepackage{bbm}
\usepackage{verbatim}
\usepackage{array}
\usepackage{booktabs}
\usepackage{tabularx}
\usepackage{url}
\usepackage{marvosym}
\usepackage[hidelinks]{hyperref} 
\hypersetup{hypertex=true,
            colorlinks=true,
            linkcolor=red,
            anchorcolor=blue!58,
            citecolor=blue!58}

\definecolor{mygray}{RGB}{188, 188, 188}
\definecolor{revise}{RGB}{0, 0, 0}
\usepackage{wrapfig}
\usepackage{colortbl}
\PassOptionsToPackage{table}{xcolor}
\usepackage[table]{xcolor}

\def\eg{\emph{e.g}.}

\def\etal{\emph{et al}.}
\def\ie{\emph{i.e}.}

\definecolor{mycolor}{RGB}{64, 128, 255}
\definecolor{lightgray}{RGB}{198, 198, 198}

\ifCLASSOPTIONcompsoc
  \usepackage[compress]{cite}
\else
  \usepackage{cite}
\fi
\ifCLASSINFOpdf
\else
\fi
\begin{document}
\title{
Frequency-aware Feature Fusion for \\ Dense Image Prediction
}
\iftrue
\author{
		Linwei~Chen,
		Ying~Fu,~\IEEEmembership{Senior Member,~IEEE},
		Lin~Gu,
		Chenggang Yan,
		Tatsuya Harada,~\IEEEmembership{Member,~IEEE},
		and
		Gao~Huang,~\IEEEmembership{Member,~IEEE},
\IEEEcompsocitemizethanks{
\IEEEcompsocthanksitem Linwei Chen and Ying Fu are with MIIT Key Laboratory of Complex-field Intelligent Sensing, Beijing Institute of Technology, Beijing, China, and School of Computer Science and Technology, Beijing Institute of Technology, Beijing, China.
\IEEEcompsocthanksitem Lin~Gu is with RIKEN AIP, Tokyo, Japan, and RCAST, The University of Tokyo, Tokyo, Japan.
\IEEEcompsocthanksitem Chenggang Yan is with the School of Automation at Hangzhou Dianzi University, Hangzhou, China.
\IEEEcompsocthanksitem Tatsuya Harada is with the Research Center for Advanced Science and Technology, The University of Tokyo, Tokyo, Japan and RIKEN AIP, Tokyo, Japan.
\IEEEcompsocthanksitem Gao Huang is with Department of Automation, Tsinghua University, Beijing, China.
}
\thanks{
$\textrm{\Letter}$ Corresponding author:
Ying Fu (fuying@bit.edu.cn)
}
}
\fi

%
%

\markboth{
}%
{Shell \MakeLowercase{\textit{et al.}}: Bare Demo of IEEEtran.cls for Computer Society Journals}
%



\IEEEtitleabstractindextext{%
\begin{abstract}
Dense image prediction tasks demand features with strong category information and precise spatial boundary details at high resolution. To achieve this, modern hierarchical models often utilize feature fusion, directly adding upsampled coarse features from deep layers and high-resolution features from lower levels. In this paper, we observe rapid variations in fused feature values within objects, resulting in intra-category inconsistency due to disturbed high-frequency features. Additionally, blurred boundaries in fused features lack accurate high frequency, leading to boundary displacement. Building upon these observations, we propose Frequency-Aware Feature Fusion (FreqFusion), integrating an Adaptive Low-Pass Filter (ALPF) generator, an offset generator, and an Adaptive High-Pass Filter (AHPF) generator. The ALPF generator predicts spatially-variant low-pass filters to attenuate high-frequency components within objects, reducing intra-class inconsistency during upsampling. The offset generator refines large inconsistent features and thin boundaries by replacing inconsistent features with more consistent ones through resampling, while the AHPF generator enhances high-frequency detailed boundary information lost during downsampling. Comprehensive visualization and quantitative analysis demonstrate that FreqFusion effectively improves feature consistency and sharpens object boundaries. Extensive experiments across various dense prediction tasks confirm its effectiveness. 
The code is made publicly available at \href{https://github.com/Linwei-Chen/FreqFusion}{https://github.com/ying-fu/FreqFusion}.

\end{abstract}

}

\maketitle

\IEEEdisplaynontitleabstractindextext

%
\IEEEpeerreviewmaketitle


%
%
%
%

 

\IEEEraisesectionheading{\section{Introduction}\label{sec:introduction}}
\IEEEPARstart{D}{ense} 
{\color{revise} image prediction encompasses various computer vision tasks that involve labeling each pixel in an image with a predefined class. 
These tasks include object detection~\cite{fasterRCNN2015}, semantic segmentation~\cite{2015fcn, deeplabv3plus}, instance segmentation~\cite{MaskRCNN2017}, and panoptic segmentation~\cite{2019panoptic}. They are crucial for scene understanding and are important for real-world applications such as autonomous driving~\cite{rashed2021generalized, zhang2016autonomous}, medical imaging~\cite{2017chestx, 2019unetnature}, and robotics~\cite{2019bonnet}. These tasks require robust category information for object classification and detailed spatial boundary information for object location.
}

\begin{figure}[t!]
\centering
\hspace{-2.918mm}
\includegraphics[width=0.9918\linewidth]{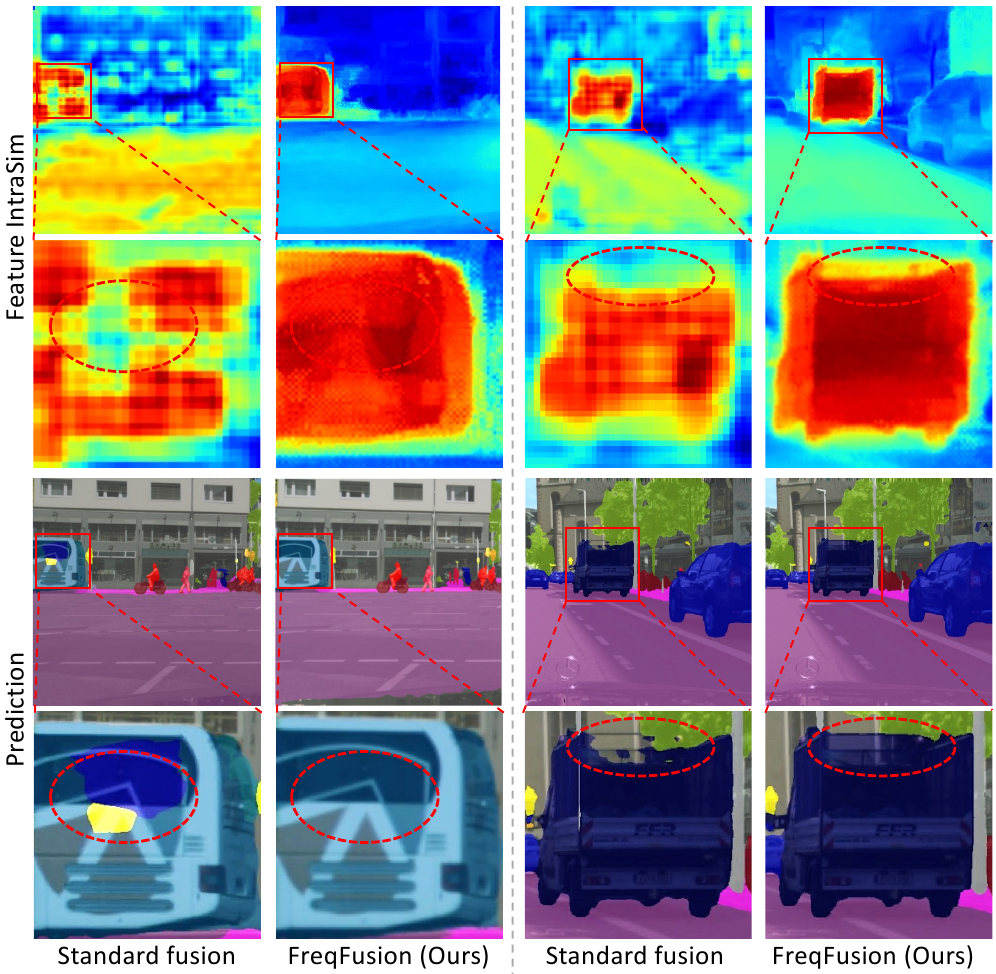}   
\vspace{-2.918mm}
\caption{
\color{revise}
Feature intra-category similarity (IntraSim) and prediction visualization. 
The brighter color indicates a higher IntraSim for the bus (left two columns) and truck (right two columns).
The standard feature fusion demonstrates low IntraSim within objects and at their boundaries. We observe rapid changes or variations in feature values within objects, \ie, disturbed high-frequency features leading to relatively low intra-category similarity~\cite{2022frequencysimilarity} and resulting in intra-category inconsistency. Furthermore, the blurred boundary lacks accurate high frequency, leading to boundary displacement.
The proposed FreqFusion shows more consistent features and clear boundaries, contributing to more consistent prediction with finer boundaries.
}
\label{fig:intro}
\end{figure}

{\color{revise}
Modern models~\cite{vgg, resnet2016, 2021swin} typically use a hierarchical design with multiple downsampling operations to progressively reduce feature size.
This process often results in the loss of detailed boundary information, which is essential for precise dense pixel-wise predictions. 
To solve this, \textit{feature fusion}~\cite{exfuse2018} is widely used~\cite{2015fcn, 2017feature, 2018upernet, 2019panopticfpn, 2019unetnature} to combine higher-level coarse features from deep layers with lower-level but high-resolution features.
Empirically, higher-level features tend to carry more category information, while lower-level features provide more boundary details~\cite{exfuse2018, 2023sfnet, 2022IFA}.
During standard feature fusion, coarse features are simply upsampled via nearest neighbor or bilinear interpolation and then added or concatenated with high-resolution features.
}

{\color{revise}
Nonetheless, standard feature fusion exhibits two issues that significantly impact dense prediction, namely, intra-category inconsistency and boundary displacement, as illustrated in Figure~\ref{fig:intro}. 
One main reason for intra-class inconsistency arises from the substantial differences between various parts of the same object~\cite{2018DFN}. For instance, the wheel of a car may exhibit more texture and darkness, while the car window appears smooth and shiny. 
But standard feature fusion~\cite{2017feature} falls short in addressing these inconsistent features. Instead, simple bilinear upsampling, commonly employed in it, may worsen the problem by upsampling a single inconsistent feature to multiple pixels, exacerbating intra-category inconsistency.
Additionally, prior studies~\cite{sapa, fade} have observed that simple interpolation often overly smooths features, resulting in boundary displacement.
}

\begin{figure}[t!]
\centering
\includegraphics[width=0.98\linewidth]{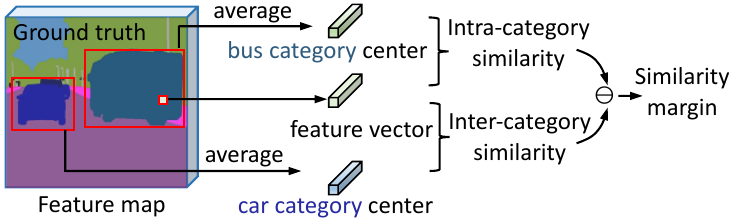}  
\vspace{-1mm}     
\caption{
Illustration of intra-category similarity, inter-category similarity, and similarity margin.
Different colors indicate different categories.
}
\label{fig:feature_similarity}
\end{figure}

To quantify these problems, we employ feature similarity analysis, as illustrated in Figure~\ref{fig:feature_similarity}. 
Intra-category inconsistency can be assessed through intra-category similarity, which measures the similarity between the feature vector and the category-wise averaged feature, \ie, category center~\cite{2023pixel2center}. 
Similarly, we can evaluate inter-class similarity, allowing us to calculate the similarity margin. 
The boundary displacement can be characterized by low intra-category similarity and similarity margin of boundary areas.
As illustrated in Figure~\ref{fig:intro}, the inconsistent features in the interior bus and truck exhibit low intra-category similarity, and the boundaries also manifest low and diminishing intra-category similarity.
Given that the classification score is determined by computing the similarity between learned category-aware fixed weights and features~\cite{2023revisiting}, features with low intra-class similarity and low similarity margin lead to low classification scores for the corresponding category and result in misclassifications.

In this paper, we observe the presence of rapid changes or variations in feature values within the object, \ie, disturbed high-frequency in the feature leads to low intra-category similarity~\cite{2022frequencysimilarity}, resulting in intra-category inconsistency. Furthermore, the blurred boundary exhibits a lack of accurate high frequency, leading to boundary displacement, as shown in Figure~\ref{fig:intro}.

{\color{revise}
Building upon these observations, we propose frequency-aware feature fusion (FreqFusion), a method designed to enhance features during the process of feature fusion. 
FreqFusion consists of three key components: Adaptive Low-Pass Filter (ALPF) generator, offset generator, and Adaptive High-Pass Filter (AHPF) generator. 
The ALPF generator predicts spatially-variant low-pass filters, aiming to reduce intra-class inconsistency by attenuating high-frequency components within objects and smoothing the features during upsampling. 
The offset generator predicts offsets to resample feature pixels and replaces features with low intra-category similarity with nearby features exhibiting high intra-category similarity, thereby refining both the interior and boundaries.
The AHPF generator extracts high-frequency details from lower-level features that cannot be recovered after downsampling, resulting in more accurate boundary delineation. These three components work collaboratively to recover fused features with consistent category information and sharp boundaries.
}

Specifically, the ALPF generator applies low-pass filters to smooth and upsample coarse high-level features, thereby reducing the disparity between pixel values and minimizing feature inconsistency. 
To prevent blurring at boundaries, inspired by~\cite{2023delving}, it predicts spatially variant low-pass filters for each upsampled feature coordinate instead of using the fixed kernel in the conventional interpolation~\cite{1999discrete}.
Through feature similarity analysis, we reveal that smooth features with spatial-variant low-pass filters can largely reduce overall feature inconsistency. 
It increases intra-category similarity and leads to a higher similarity margin, thereby enhancing the consistency and discriminative nature of the features. Consequently, it benefits dense prediction tasks.

While the use of smooth features with the ALPF generator increases overall intra-category similarity, it may not excel in rectifying large areas of inconsistent features or fine boundaries. 
Expanding the size of low-pass filters is beneficial for addressing large areas of inconsistent features but can be detrimental to thin and boundary areas. Conversely, reducing the size of low-pass filters benefits thin and boundary areas but hinders the correction of large areas with inconsistent features
To address this contradiction, we introduce the offset generator. 
It is motivated by the observation that low intra-category similarity features often have neighbors with high intra-category similarity, as shown in Figure~\ref{fig:intro}. 
The offset generator first calculates local similarity and then predicts an offset in the direction of high similarity for resampling. 
This approach allows for resampling features with high intra-category similarity to replace features with low intra-category similarity. 
Thus, the offset generator can rectify inconsistent features in both large areas and thin boundary regions.

Although the ALPF generator and offset generator effectively recover upsampled high-level features with high intra-class consistency and refined boundaries, the detailed boundary information in lower-level features lost during downsampling cannot be fully restored in high-level features.
According to the Nyquist-Shannon Sampling Theorem \cite{shannon1949communication, nyquist1928}, frequencies higher than the Nyquist frequency, which is equivalent to half of the sampling rate, are permanently lost during downsampling. For instance, frequencies above $\frac{1}{4}$ become aliased during a 2$\times$ downsampling operation (\eg, a 1$\times$1 convolution layer with a stride of 2 has a sampling rate of $\frac{1}{2}$).
To address this limitation, we introduce the AHPF generator, which extracts detailed boundary information by predicting and applying spatially variant high-pass filters to low-level features, thereby enhancing the high-frequency power above the Nyquist frequency and sharping the boundary. 
Frequency analysis demonstrates an improvement in high-frequency power, resulting in finer visualized dense prediction results.

Qualitative results showcase the effectiveness of FreqFusion in recovering high-resolution features with discriminative category information and clear boundaries. 
Quantitative analysis reveals significant improvements in intra-category similarity and similarity margin.
This, in turn, leads to substantial performance enhancements across various tasks, including semantic segmentation, object detection, instance segmentation, and panoptic segmentation, outperforming previous state-of-the-art methods.
Specifically,
1) for {\it semantic segmentation}, FreqFusion enhances SegFormer-B1~\cite{2021segformer} and SegNeXt-T~\cite{segnext} by 2.8 and 2.0 mIoU, respectively. It achieves a gain of +1.4/0.7 mIoU for Heavy Mask2Former-Swin-B/L~\cite{2022mask2former} on the Challenging ADE20K dataset~\cite{ade20k}.
2) for {\it object detection}, FreqFusion boosts AP by 1.8 with Faster R-CNN-R50~\cite{fasterRCNN2015} on MS COCO~\cite{2014microsoft};
3) for {\it instance segmentation}, FreqFusion improves the performance of Mask R-CNN-R50~\cite{MaskRCNN2017} by 1.7 box AP and 1.3 mask AP on MS COCO~\cite{fasterRCNN2015};
4) for {\it Panoptic Segmentation}, FreqFusion outperforms other upsamplers by significant margins, achieves a 1.9 PQ improvement with Panoptic FPN-R50~\cite{2019panopticfpn} on MS COCO~\cite{fasterRCNN2015}.

Our main contributions can be summarized as follows:
\begin{itemize}
\item We identify two significant issues present in widely-used standard feature fusion techniques: intra-category inconsistency and boundary displacement. We also introduce feature similarity analysis to quantitatively measure these issues, which not only contributes to the development of new feature fusion methods but also has the potential to inspire advancements in related areas and beyond.
\item We propose FreqFusion, which addresses category inconsistency and boundary displacement by adaptively smoothing the high-level feature with spatial-variant low-pass filters, resampling nearby category-consistent features to replace inconsistent features in the high-level feature, and enhancing the high frequency of lower-level features.
\item Qualitative and quantitative results demonstrate that FreqFusion increases intra-category similarity and similarity margin, leading to a consistent and considerable improvement across various tasks, including semantic segmentation, object detection, instance segmentation, and panoptic segmentation.
\end{itemize}

{
\color{revise}
The paper is structured as follows: 
Section~\ref{sec:related_work} introduces related work. 
Section~\ref{sec:metric} describes the similarity metric for analysis.
Section~\ref{sec:method} presents the proposed method, accompanied by extensive visual and quantitative feature analysis results demonstrating its effectiveness in addressing feature inconsistency and boundary displacement.
Section~\ref{sec:experiments} showcases experimental results. Section~\ref{sec:discussion} outlines the significant differences between the proposed method and previous approaches. Finally, Section~\ref{sec:conclusion} summarizes the paper.
}

\section{Related Work}
\label{sec:related_work}
We begin by reviewing dense prediction tasks and techniques for feature fusion and aggregation. Subsequently, we introduce advancements in frequency domain learning.

\subsection{Dense Image Prediction.} 
Dense image prediction tasks encompass various challenges, such as object detection~\cite{fasterRCNN2015, 2021crafting}, semantic segmentation~\cite{2015fcn, deeplabv3plus, chen2023semantic, 2023casid}, instance segmentation~\cite{MaskRCNN2017, 2023lis, 2021efficienthybrid, 2022hybridsupervised}, and panoptic segmentation~\cite{2019panoptic}.
The advancements in dense prediction primarily hinge on a few seminal deep network architectures. 
For instance, since the advent of fully convolutional networks (FCNs) in semantic segmentation~\cite{2015fcn}, the field has evolved with foundational segmentation architectures like U-Net~\cite{2019unetnature}, SegNet~\cite{2017segnet}, and DeepLab\cite{2017deeplab, deeplabv3plus}. 
Similarly, in object detection, models such as R-CNN\cite{fasterRCNN2015} and YOLO~\cite{2016yolo} have dominated the landscape in recent years. 
Subsequently, typical network architectures like Feature Pyramid Networks (FPNs)~\cite{2017feature} have been widely used in other dense prediction tasks, including instance segmentation and panoptic segmentation.

Different from image processing or generation tasks~\cite{chen2022consistency, 2022levelAware, 2022gan, 2022Guided, fuying-2021-tpami, wei2021physics, liu2024transformer, li2024supervise, zhang2024deep, 10406185, 10360444, 10231043}, in these representative architectures for dense prediction, feature fusion is an essential component. This is because most backbone architectures involve multiple downsampling stages~\cite{resnet2016, 2021swin, 2022convnet}, while the expected output often requires high-resolution information for accurate object classification and detailed spatial boundaries for precise object localization.
Feature fusion offers a simple solution by combining low-level, high-resolution features with coarse, high-level features. In this work, we further propose an effective feature fusion method named FreqFusion, to obtain high-quality fused features with consistent category information and clear boundaries. The proposed FreqFusion seamlessly integrates with state-of-the-art architectures, from CNNs (\eg, SegNeXt~\cite{segnext}) to Transformers (\eg, SegFormer~\cite{2021segformer}), providing a performance boost with minimal additional parameters and computational cost.

We select a few representative models as baselines in different tasks, including SegNeXt~\cite{segnext}, SegFormer~\cite{2021segformer}, and Mask2Former for semantic segmentation, Faster R-CNN for object detection, Mask R-CNN~\cite{MaskRCNN2017} for instance segmentation, Panoptic FPN~\cite{2019panoptic} for panoptic segmentation). We demonstrate how FreqFusion is applied to these tasks and how it improves upon these baselines.




\subsection{Feature Fusion and Aggregation.}
Feature fusion is a process to fuse lower-level high-resolution features and higher-level coarse features to obtain details and semantic information~\cite{exfuse2018}.
While feature aggregation aggregates features from different network stages, \ie, it consists of a series of feature fusion operations.

There are two most common feature aggregation architectures, one is top-down, \eg, SegNet~\cite{2017segnet}, U-Net~\cite{2019unetnature}, RefineNet~\cite{refinenet2017}, and DeepLabv3+~\cite{deeplabv3plus}, which fuse features from low resolution to high resolution. 
The other is bottom-up, \eg, FRRN~\cite{2017FRRN}, DLA~\cite{2018DLA}, and SeENet~\cite{2019SeENet}, which aggregate features from high resolution to low resolution. 
Owing to the gap between resolution and semantic levels, a simple fusion strategy is less effective~\cite{exfuse2018}. 

Next, we introduce recent methods to improve feature fusion, which can be categorized into two types: 
1) kernel-based, and 2) sampling-based.

\vspace{+1mm}
\noindent{\bf Kernel-based.}
To fuse low and high-resolution features, feature upsampling is needed to upscale the low-resolution features.
Traditional upsampling operations, \eg, nearest neighbor and bilinear interpolation use fixed/hand-crafted kernels that are defined by the relative distance between pixels.
Though kernel parameters of deconvolution~\cite{2014visualizing}, Pixel shuffle~\cite{2016pixelshuffle}, and DUpsampling~\cite{2019dupsample} are learnable, the upsampling kernels are fixed and spatially-invariant once learned.
The importance of dynamic property has been emphasized recently.
As a hand-crafted operator, un-pooling~\cite{2017segnet} has dynamic upsampling behavior, in other words, upsampled positions are conditioned on max pooling indices.
CARAFE~\cite{carafe++} dynamically reassembles the local decoder features in a context-aware manner.
Similar to CARAFE, IndexNet~\cite{2020indexnet} and A2U~\cite{2021a2u} also only consider the assets of encoder features for dynamic upsampling.
SAPA~\cite{sapa} and FADE~\cite{fade} further consider the assets of both encoder and decoder features for generating upsampling kernels.
In summary, the essence of these operators lies in the data-dependent upsampling kernels, whose parameters are predicted by a sub-network. This underscores a promising avenue for exploring better feature upsampling.


\vspace{+1mm}
\noindent{\bf Sampling-based.}
Recently, a series of methods aim to improve feature fusion by adjusting sampling coordinates.
GUM~\cite{2018gun} learns the guidance offsets and applies those offsets to upsampled feature maps.
SFNet~\cite{2023sfnet} warps coarse high-level features with predicted offsets for alignment.
AlignSeg~\cite{2021alignseg} further predicts offsets for aligning multi-resolution features and context modeling.
FaPN~\cite{2021fapn} utilizes deformable convolution~\cite{2017deformable} to align features from the coarsest resolution to the finest resolution progressively.
IFA~\cite{2022IFA} aligns multi-resolution features with implicit offsets by using implicit neural representation.
Dysample~\cite{2023dysample} learn to upsample by learning to sample the coarse high-level feature dynamically.
These methods apply explicit or implicit spatial offsets to align the low and high-resolution features.
Moreover, recent works~\cite{2018DFN, exfuse2018, gatedscnn2019} employ channel attention or gates to combine lower and higher-level features, using adaptive weights conditioned on higher-level features rather than equal weights.
The Mask2Former utilizes a deformable attention module~\cite{deformableattention}, which applies both spatial offsets and adaptive attention weights to fuse multi-scale features.

Our work is closely related to both kernel-based and dynamic sampling-based methods. While previous studies empirically observe the problem in standard feature fusion, they lack clear definitions supported by quantitative measurements. 
In contrast, we clearly identify and define the issues of intra-category inconsistency and boundary displacement, measuring them through feature similarity analysis. 
The proposed FreqFusion effectively addresses these issues with the goal of achieving simultaneous feature consistency and boundary sharpness.

\begin{figure*}[t!]
\centering
\color{revise}
\includegraphics[width=0.8918\linewidth]{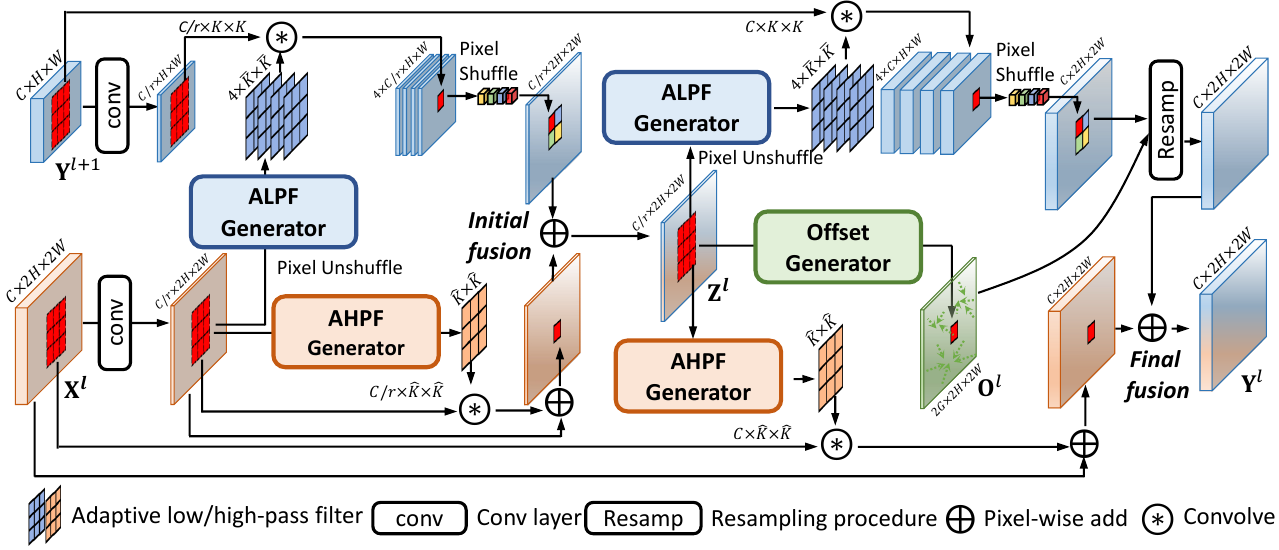}    
\vspace{-1.918mm}     
\caption{
The illustration of FreqFusion.
Pixel unshuffle involves resizing the spatial dimensions of the feature by half and expanding the channel by a factor of 4$\times$, dividing them into 4 groups, such as from $C\times 2H \times 2W$ to $4\times C\times H\times W$. Pixel shuffle~\cite{2016pixelshuffle} is the reverse operation, transitioning from $4\times C\times H\times W$ to $C\times 2H \times 2W$.
The Adaptive Low-Pass Filter (ALPF) generator and Adaptive High-Pass in the initial fusion share the same parameters as those in the final fusion.
}
\label{fig:freqfusion}
\end{figure*}
\begin{figure*}[t!]
\color{revise}
\centering
\includegraphics[width=0.9518\linewidth]{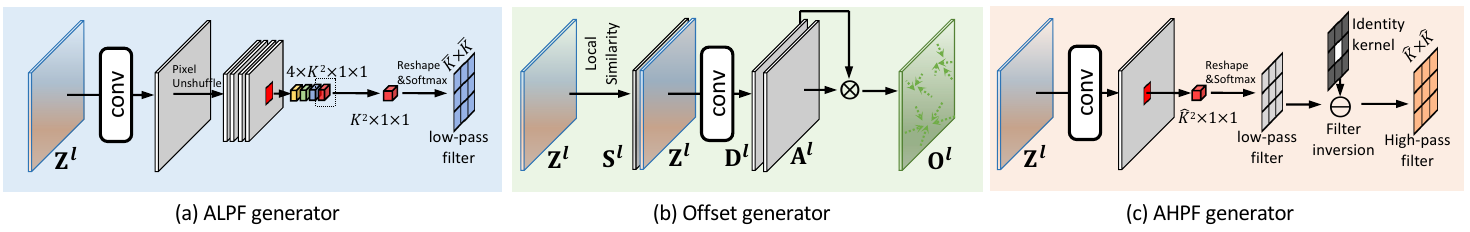}    
\vspace{-1.918mm}     
\caption{
The illustration of generators in FreqFusion. $\otimes$ represents element-wise multiplication, and $\ominus$ represents subtraction.
}
\label{fig:generators}
\end{figure*}

\subsection{Frequency Domain Learning}
Frequency-domain analysis, as a fundamental tool~\cite{1994digital, 2000digital}, has long been proven to be an effective tool for traditional signal processing.
Recently, a series of works have introduced frequency analysis to deep learning. 

In this context, they are employed to examine the optimization strategies~\cite{2019fourier} and generalization capabilities~\cite{2020highfrequency} of Deep Neural Networks (DNNs). 
Rahaman~\etal~\cite{2019spectralbias} and Xu~\etal~\cite{2021deepfrequencyprinciple} find the effective target function for a deeper hidden layer bias towards lower frequencies during training, thus these networks prioritize learning the low-frequency modes, this phenomenon is called spectral bias/frequency principle.
Zhang~\etal~\cite{2019makingshiftinvariant} investigate how frequency aliasing impacts the shift-invariance of modern models, and subsequently, AdaBlur~\cite{2023delving} applied content-aware low-pass filters during downsampling for anti-aliasing. Additionally, FLC~\cite{2022flc} also demonstrated that frequency aliasing degrades the robustness of models. Qin~\etal~\cite{2021fcanet} and Magid~\etal~\cite{2021dynamic} explore utilizing more frequency components obtained from discrete cosine transform coefficients for channel attention mechanisms.
Huang~\etal~\cite{2023adaptivefrequency} employ the conventional convolution theorem in DNNs, demonstrating that adaptive frequency filters can efficiently serve as global token mixers. 
A series of frequency-domain techniques have also been seamlessly integrated into DNN architectures, facilitating the learning of non-local features\cite{2020ffc, 2021gfnet, 2021FNO, 2022AFNO, 2023adaptivefrequency}.
Chen~\etal~\cite{2023lis} demonstrate how the low-pass filter suppress high frequency feature noise caused by noise in the images,  effectively addressing the challenges of instance segmentation in low-light conditions.
Many works demonstrate adversarial attack can be achieved by manipulate the high frequency components~\cite{2019fourier, 2022frequencysimilarity, 2022exploringfrequency}
Luo~\etal~\cite{2022frequencysimilarity} demonstrate that perturbing high frequencies leads to a large reduction in intra-category similarity, thereby degrading feature representations.

In this work, we consider intra-category inconsistency as the presence of disturbed high frequency, significantly reducing intra-category similarity, as observed in~\cite{2022frequencysimilarity}. Boundary displacement is regarded as a lack of high frequency, as noted in~\cite{2021dynamic}. Thus, we employ adaptive low-pass filters to reduce feature inconsistency and high-pass filters to enhance useful high-frequency details and sharpen boundaries. This demonstrates an innovative application of frequency-domain techniques in addressing intra-category inconsistency and boundary displacement in feature fusion, benefiting various fundamental computer vision tasks.

{\color{revise}
\section{Feature Similarity Analysis Metrics}
\label{sec:metric}
We begin by introducing metrics for feature similarity analysis. These metrics aim to quantify both intra-category inconsistency and boundary displacement issues that emerge during the process of feature fusion. This establishes a solid foundation for developing and analyzing effective feature fusion techniques.

Feature similarity is widely used for assessing the quality of extracted features~\cite{2023pixel2center, 2023generalized, 2022maximizing, 2023semanticsimilarity}. 
Typically, features within the same category should show high similarity, ensuring high intra-category similarity. On the other hand, features from different categories should exhibit low similarity, resulting in low inter-category similarity.
A large gap between intra-category and inter-category similarities, referred to as similarity margin, is crucial for preventing misclassification.

To facilitate a quantitative assessment of intra-category inconsistency and boundary displacement, as well as to assess the quality of the fused features, we introduce metrics encompassing intra-category similarity, similarity margin, and similarity accuracy. These metrics offer a comprehensive framework for evaluating the discriminative power of extracted feature maps.

\vspace{+1mm}
\noindent{\bf Intra-category \& inter-category similarity.}
Intra-category similarity is computed by first deriving the category center through the averaging of features within each category. Subsequently, we calculate the cosine similarity between the category center and the feature vector belonging to the same category. This is expressed as:
\begin{equation}
\footnotesize
\begin{aligned}
\text{IntraSim}(\mathbf{Y}_{i,j}^{cls=1}) = \text{CosSim}(\mathbf{Y}_{i,j}^{cls=1}, \frac{1}{|\Omega^{cls=1}|} \sum_{i,j \in \Omega^{cls=1}}\mathbf{Y}_{i,j}).
\end{aligned}
\end{equation}
Here, we consider binary segmentation that has two categories, $cls=1$ denotes the ground truth category of feature vector $\mathbf{Y}_{i,j}$, $\Omega^{cls=1}$ represents the area belonging to category 1, and $\text{CosSim}$ is the cosine similarity.
Similarly, the inter-category similarity is calculated using the same method, with the distinction that the category center and feature vector are from different categories.
\begin{equation}
\footnotesize
\begin{aligned}
\text{InterSim}(\mathbf{Y}_{i,j}^{cls=1}) = \text{CosSim}(\mathbf{Y}_{i,j}^{cls=1}, \frac{1}{|\Omega^{cls=0}|} \sum_{i,j \in{\Omega}^{cls=0}}\mathbf{Y}_{i,j}).
\end{aligned}
\end{equation}
Here, $\Omega^{cls=0}$ indicates the area that belong to category from $\mathbf{Y}_{i,j}$.

\vspace{+1mm}
\noindent{\bf Similarity margin.}
Consequently, the similarity margin is determined by subtracting the inter-category similarity from the intra-category similarity
\begin{equation}
\begin{aligned}
\text{SimMargin}(\mathbf{Y}_{i,j}) = \text{IntraSim}(\mathbf{Y}_{i,j}) - \text{InterSim}(\mathbf{Y}_{i,j}).
\end{aligned}
\end{equation}

\vspace{+1mm}
\noindent{\bf Similarity accuracy.}
To comprehensively assess the risk of misclassification rate resulting from intra-category inconsistency and boundary displacements, we assign each feature to a category based on its most similar category center. Thus, we can calculate the accuracy, \ie, similarity accuracy.
It measures the proportion of features exhibiting greater inter-category similarity than intra-category similarity across all categories.

The metrics of intra-category similarity, similarity margin, and similarity accuracy collectively evaluate the category information present in the features, which provides discriminative power and separation between different categories.
}

{\color{revise}
\section{Frequency-aware Feature Fusion}
\label{sec:method}
In this section, we introduce FreqFusion as shown in Figure~\ref{fig:freqfusion}. It consists of three essential components: the Adaptive Low-Pass Filter (ALPF) generator, the offset generator, and the Adaptive High-Pass Filter (AHPF) generator, as illustrated in Figure~\ref{fig:generators}.

FreqFusion operates through two primary stages, \ie, initial fusion and final fusion. 
Prior to the final fusion step, it is necessary to compress and fuse both low-level and high-level features to serve as input for the three generators, ensuring efficiency in the final fusion stage.
We first introduce how we enhance initial fusion, elucidating its significance within the FreqFusion framework. Subsequently, we provide detailed insights into the functioning of each of the three generators, thereby offering a comprehensive understanding of their roles in the fusion process.
}

{\color{revise}
\subsection{Overview of FreqFusion}
We begin by presenting the widely-used standard feature fusion approach, followed by an overview of the design of FreqFusion.

\vspace{+1mm}
\noindent{\bf Standard feature fusion.}
Generally, a common way of feature fusion can be formulated as~\cite{2016fcn, exfuse2018, 2017feature}:
\begin{equation}
\begin{aligned}
\label{eq:fusion}
\mathbf{Y}^{l} = \mathcal{F}^{\text{UP}}(\mathbf{Y}^{l+1}) + \mathbf{X}^{l},
\end{aligned}
\end{equation}
where $\mathbf{X}^{l} \in \mathbb{R}^{C\times 2H \times 2W}$, $\mathbf{Y}^{l+1} \in \mathbb{R}^{C\times H \times W}$ represent the $l$-th features generated by the backbone and the fused feature at the $l$-th level, respectively. 
We assume they have the same number of channels; if not, a simple projection function like a $1\times 1$ convolution can ensure this~\cite{2017feature}, which we omit for brevity. 
The term $\mathcal{F}^{\text{UP}}$ denotes upsampling, for example, $2\times$ nearest neighbor or bilinear interpolation~\cite{2017feature, 2018upernet}.

Although widely used, this straightforward approach to feature fusion manifests two issues that detrimentally impact dense prediction, \ie, intra-category inconsistency and boundary displacement.
Standard fusion falls short in rectifying these inconsistent features, and simple interpolation within it may even worsen the problem by upscaling a single inconsistent feature to multiple inconsistent pixels.
Furthermore, as observed in various prior works~\cite{2020indexnet, sapa, 2015fcn}, outputs from simple interpolation often lean towards excessive smoothness, resulting in boundary displacement.
Additionally, the detailed boundary information in lower-level features is not fully utilized.
}

\begin{figure}[t!]
\centering
\scalebox{0.9918}{
\begin{tabular}{cccccc}
\hspace{-2.918mm}
\includegraphics[width=0.9918\linewidth]{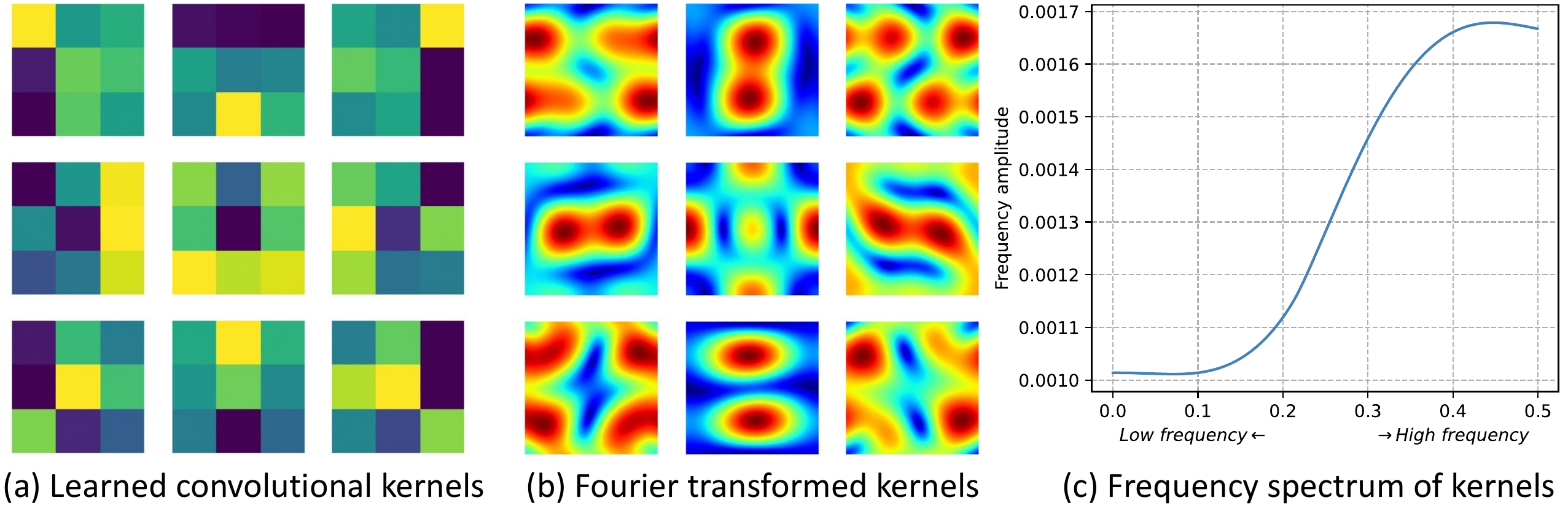} 
\end{tabular}
}
\caption{
Frequency analysis of the learned convolutional kernel in the ALHPF generator is presented. In (a), the nine learned kernels for generating $3\times 3$ spatial-variant low-pass filters are displayed. A brighter color indicates a higher learned weight. (b) illustrates their corresponding Fourier-transformed kernels. To further analyze their characteristics, we average their frequency amplitudes and present the frequency spectrum in (c), demonstrating higher power for high-frequency components, indicating reliance on high-frequency components in the feature for filter prediction.
}
\label{fig:conv_frequency_analysis}
\end{figure}

\begin{figure}[t!]
\color{revise}
\centering
\scalebox{0.9918}{
\begin{tabular}{cccccc}
\hspace{-2.918mm}
\includegraphics[width=0.9918\linewidth]{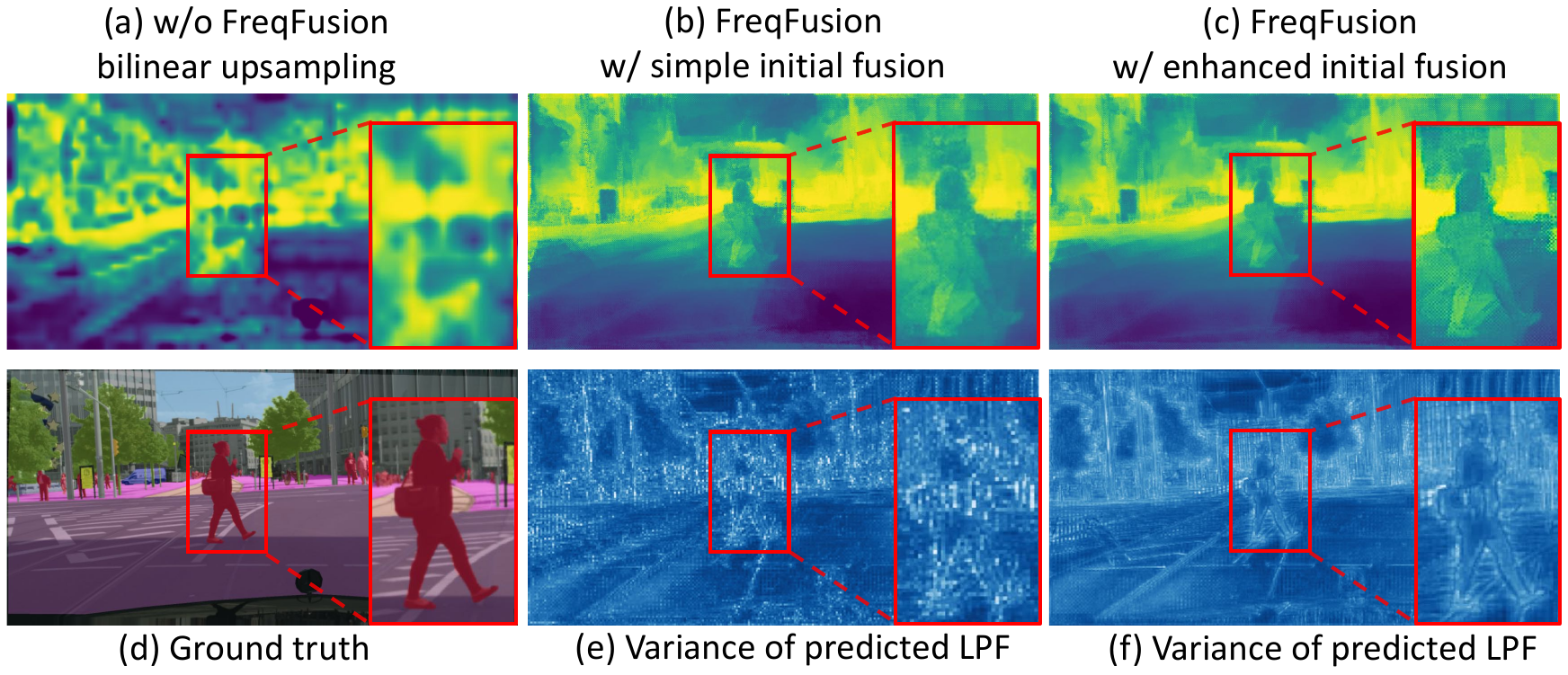} 
\end{tabular}
}
\caption{
Visualization for initial fusion.
In comparison with (a), the FreqFusion recovers more detailed features in (b). Moreover, with enhanced initial fusion, (c) exhibits clearer boundaries.
(e) and (f) depict the standard variance of predicted low-pass filters (LPF), (f) demonstrates a more effective maintenance of boundaries.
}
\label{fig:init_fusion}
\end{figure}

{\color{revise}
\vspace{+1mm}
\noindent{\bf Design of FreqFusion.}
As shown in Figure~\ref{fig:freqfusion}, the proposed FreqFusion can be formally written as:
\begin{equation}
\begin{aligned}
\mathbf{Y}^{l}_{i,j} &= \mathbf{\tilde Y}^{l+1}_{i+u,j+v} + \mathbf{\tilde X}^{l}_{i,j}, \\
\mathbf{\tilde Y}^{l+1} &= \mathcal{F}^{\text{UP}}(\mathcal{F}^{\text{LP}}(\mathbf{Y}^{l+1})),
\quad
\mathbf{\tilde X}^{l} = \mathcal{F}^{\text{HP}}(\mathbf{X}^{l}) + \mathbf{X}^{l}.
\end{aligned}
\end{equation}
where $\mathcal{F}^{\text{LP}}$ denotes the low-pass filters predicted by the ALPF generator,
$(u, v)$ indicates the offset values predicted by the offset generator for the feature coordinates at $(i, j)$,
and $\mathcal{F}^{\text{HP}}$ represents the high-pass filters predicted by the AHPF generator, respectively.
They address category inconsistency and boundary displacement by adaptively smoothing the high-level feature with spatial-variant low-pass filters, resampling nearby category-consistent features to replace inconsistent features in the high-level feature, and enhancing the high-frequency boundary details of lower-level features.

To efficiently generate the low-pass filters $\mathcal{F}^{\text{LP}}$, offset values $(u, v)$, and high-pass filter $\mathcal{F}^{\text{HP}}$, it is necessary to first compress $\mathbf{X}^{l}$ and $\mathbf{Y}^{l+1}$ and fuse them for input into the three generators, a process we refer to as \textit{initial fusion}.
A simple initial fusion~\cite{fade, 2023sfnet, 2021alignseg} can be formally expressed as:
\begin{equation}
\label{eq:init_fusion}
\begin{aligned}
\mathbf{Z}^{l} = \mathcal{F}^{\text{UP}}(\text{Conv}_{1\times 1}(\mathbf{Y}^{l+1}))) + \text{Conv}_{1\times 1}(\mathbf{X}^{l}). \\
\end{aligned}
\end{equation}
where $\mathbf{Z}^{l} \in \mathbb{R}^{C/r\times 2H \times 2W}$ indicates the fused compressed feature, and the $r$ is the channel reduction rate for reducing the following computational cost of three generators.
The $1\times 1$ convolutional layer is utilized for channel compression.
Next, we proceed to explain how we enhance the initial fusion, followed by describing the details of the three generators
}

{\color{revise}
\subsection{Enhancing Initial Fusion}
The three generators rely on the initially fused compressed feature $\mathbf{Z}^{l}$ to predict adaptive filters and resampling offsets. 
However, the simple initial fusion presented in Equation~{\color{red}\eqref{eq:init_fusion}} exhibits two suboptimal aspects, which can adversely affect the subsequent three generators. 
Firstly, it employs simple interpolation for upsampling the compressed feature, resulting in blurred boundaries~\cite{sapa, 2023dysample}. Secondly, frequency analysis reveals that the ALPF generator heavily relies on high-frequency information in the fused compressed feature. 
However, traditional convolutional layers can only capture fixed patterns of high frequency. Therefore, we propose further enhancements to the initial fusion process.

\vspace{+1mm}
\noindent{\bf Upsampling for initial fusion.}
Several seminal works~\cite{sapa, fade, 2023sfnet} have emphasized the significance of upsampling, highlighting that simple interpolation techniques, such as nearest-neighbor or bilinear interpolation, can introduce smooth and inaccurate boundaries. 
Despite this awareness, for generating the initial upsampled intermediate feature, they~\cite{fade, 2023sfnet} still employ simple interpolation, inevitably leading to similar issues in the intermediate feature. Consequently, this results in suboptimal upsampling outcomes in subsequent stages.

To address this issue, as depicted in Figure~\ref{fig:freqfusion}, we utilize the ALPF generator to take compressed low-level features as input and generate an initial low-pass filter to upsample the compressed high-level features. Leveraging the high-resolution structure present in the low level can be beneficial for upsampling coarse high-level features~\cite{fade}. 
The detail of ALPF generator is described in Section~\ref{sec:ALPF}.
By adopting this approach, we circumvent the use of simple interpolation, resulting in finer initial fusion results and benefiting subsequent generators.

\vspace{+1mm}
\noindent{\bf High-frequency enhancement for initial fusion.}
The frequency analysis, as illustrated in Figure~\ref{fig:conv_frequency_analysis}, highlights a distinct reliance of the ALPF generator on high-frequency information within the fused compressed feature. Notably, this reliance stems from the inherent nature of convolutional layers, which are constrained to capturing fixed patterns of high frequency.

Building upon this insight, we propose employing the AHPF generator as a strategic enhancement. The details of the AHPF generator are described in Section~\ref{sec:AHPF}.
The AHPF generator, as a dynamic component in our framework, is crafted to extract high-frequency components from the feature map, thus overcoming limitations posed by standard convolutional layers. Unlike convolutions with fixed learned weights, the spatially variant high-pass filter utilized by the AHPF generator demonstrates an adaptive capability to capture high-frequency patterns.

Consequently, the AHPF generator enriches the feature representation with finely tuned high-frequency details, thereby facilitating more effective downstream processing.
As depicted in Figure~\ref{fig:init_fusion}, the enhanced initial fusion benefits the subsequent generators to better adapt to the feature content, resulting in finer final fused results.
Quantitative feature similarity analysis in Table~\ref{tab:sim_analysis} also demonstrates the benefits of the ALPF and AHPF generators in improving feature consistency and boundary sharpness.
}

\begin{figure}[t!]
\centering
\includegraphics[width=0.9918\linewidth]{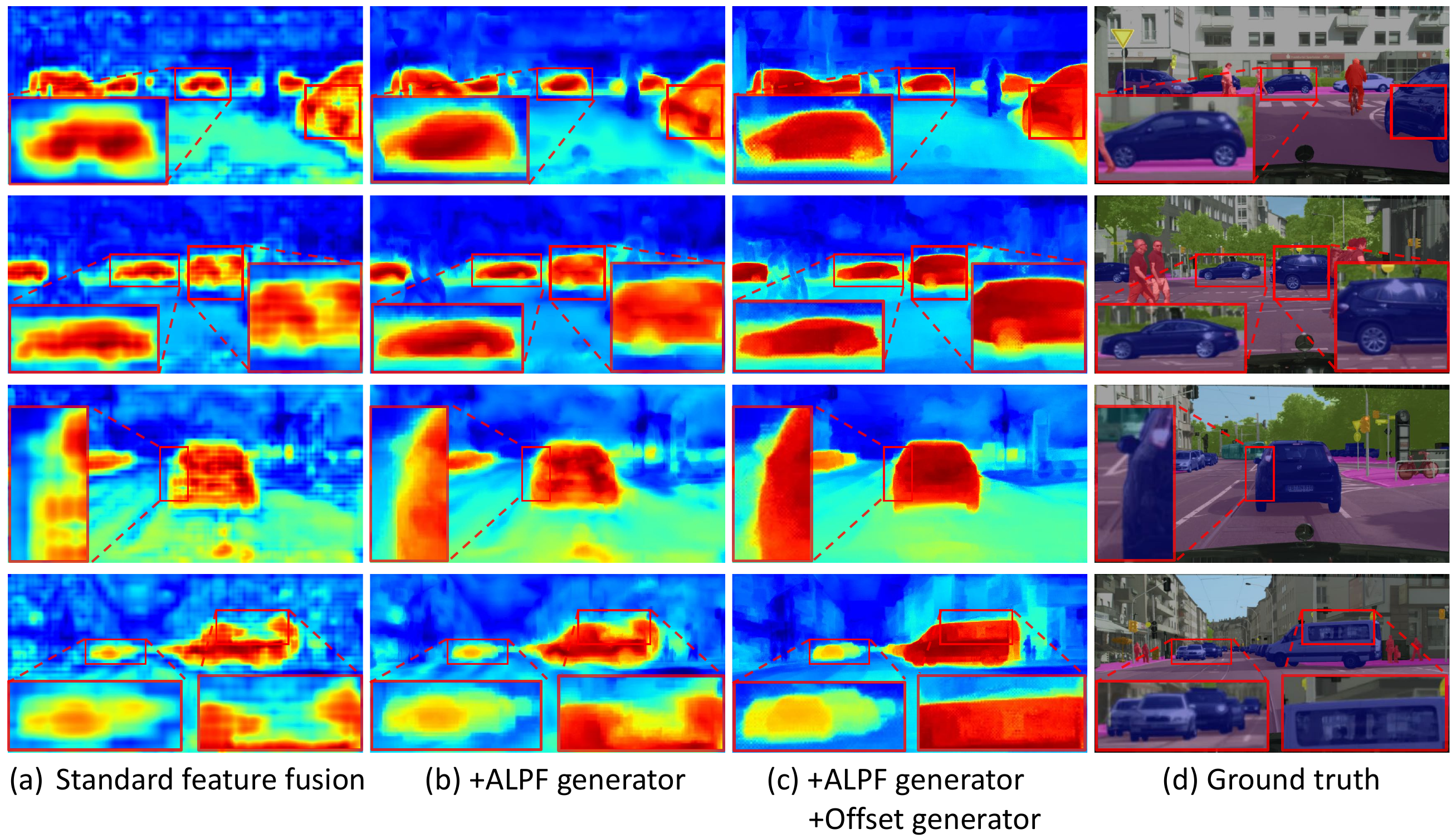} 
\caption{
Visualization for intra-category similarity of features.
We illustrate the intra-category similarity for the car category, and a brighter color indicates a higher similarity.
The bilinear upsampled results exhibit severe inconsistencies in features and displacements along the boundaries. Our proposed ALPF generator addresses these inconsistencies, enhancing feature consistency and refining boundaries. Furthermore, the offset generator not only further improves feature consistency but also contributes to achieving more accurate boundary delineations.
}
\label{fig:feat_sim_analysis}
\end{figure}

\begin{table}[tb!]
\centering
\caption{
Quantitative feature analysis results.
FreqFusion considerably improves intra-category similarity (IntraSim), similarity margin (SimMargin), and similarity accuracy (SimAcc) for both overall and boundary regions, alleviating intra-category inconsistency and boundary displacement.
}
\scalebox{0.5918}{
\begin{tabular}{l|ccc|ccccc}
\toprule[1.28pt]
\multirow{2}{*}{Method}& \multirow{2}{*}{IntraSim$\uparrow$} & \multirow{2}{*}{SimMargin$\uparrow$} & \multirow{2}{*}{SimAcc$\uparrow$} & \multicolumn{3}{c}{Boundary} \\
& & & &  IntraSim$\uparrow$ & SimMargin$\uparrow$ & SimAcc$\uparrow$ \\
\midrule
Standard feature fusion & 0.697 & 0.255 & 0.907 & 0.584 & 0.202 & 0.632 \\
\midrule
\rowcolor{mygray!58}
FreqFusion &\bf 0.799 &\bf 0.297 &\bf 0.941 &\bf 0.694 &\bf 0.239 &\bf 0.728 \\
w/o ALPF Generator (Initial fusion) & 0.792 & 0.288 & 0.929 & 0.687 & 0.230 & 0.725 \\
w/o AHPF Generator (Initial fusion) & 0.798 & 0.296 & 0.940 & 0.690 & 0.232 & 0.726 \\
w/o ALPF Generator (Final fusion) & 0.727 & 0.245 & 0.918 & 0.610 & 0.229 & 0.724 \\
w/o Offset Generator (Final fusion) & 0.760 & 0.295 & 0.925 & 0.648 & 0.235 & 0.720 \\
w/o AHPF Generator (Final fusion) & 0.796 & 0.295 & 0.938 & 0.688 & 0.228 & 0.718 \\
\bottomrule[1.28pt]
\end{tabular}
}
\label{tab:sim_analysis}
\end{table}

{\color{revise}
\subsection{Adaptive Low-Pass Filter Generator}
\label{sec:ALPF}
The Adaptive Low-Pass Filter (ALPF) generator is designed to predict dynamic low-pass filters, aiming to effectively smooth high-level features to mitigate feature inconsistency~\cite{2022frequencysimilarity} and subsequently upsample the high-level feature.
To achieve high-quality adaptive low-pass filters, it is crucial to leverage the advantages of both high-level and low-level features~\cite{fade}.
Thus, the ALPF generator takes the initially fused $\mathbf{Z}^{l}$ as input and predicts spatial-variant low-pass filters. 
It comprises a $3\times 3$ convolutional layer followed by a softmax layer, which is represented as:
\begin{equation}
\begin{aligned}
\mathbf{\bar V}^{l} &= \text{Conv}_{3\times 3}(\mathbf{Z}^{l}), \\
\mathbf{\bar W}_{i, j}^{l,p,q} &=\text{Softmax}(\mathbf{\bar V}^{l}_{i, j}) = \frac{\exp(\mathbf{\bar V}_{i, j}^{l,p,q})}{\sum_{p,q\in \Omega} \exp(\mathbf{\bar V}_{i, j}^{l,p,q})}, \\
\end{aligned}
\end{equation}
where $\mathbf{\bar V}^{l} \in \mathbb{R}^{\bar K^{2}\times 2H \times 2W}$ represents spatially-variant filter weights, where $\bar K$ indicates the kernel size of the low-pass filter.
After reshaping, $\mathbf{\bar V}^{l}$ contains $\bar K \times \bar K$ filters for each position.
Here, $\Omega$ denotes a size of $\bar K\times \bar K$.
Upon passing through a kernel-wise softmax to constrain the filters to be all positive and sum to one, the results are smooth and low-pass filters in $\mathbf{\bar W} \in \mathbb{R}^{\bar K^{2}\times 2H \times 2W}$~\cite{2023delving}.

Next, we upscale $\mathbf{Y}^{l+1} \in \mathbb{R}^{C\times H \times W}$ using a sub-pixel upsampling technique~\cite{2016pixelshuffle}. 
Specifically, we reshape $\mathbf{\bar W}^l$ in a pixel unshuffle way~\cite{2016pixelshuffle}, reducing the height and width by half and expanding the channel by 4$\times$. We then divide the channels into 4 groups, with each group having a spatially-variant low-pass filter denoted as $\mathbf{\bar W}^{l,g} \in \mathbb{R}^{\bar K^{2}\times H \times W}$, where $g\in\{1,2,3,4\}$ indicates the group. 
Consequently, we obtain 4 groups of low-pass filtered features, represented as $\mathbf{\tilde Y}^{l+1, g} \in \mathbb{R}^{C\times H \times W}$, which are then rearranged to form a 2$\times$ upsampled feature $\mathbf{\tilde Y}^{l+1} \in \mathbb{R}^{C\times 2H \times 2W}$ as:
\begin{equation}
\begin{aligned}
\mathbf{\tilde Y}^{l+1, g}_{i, j} &=  \sum_{p,q\in \Omega} \mathbf{\bar W}_{i, j}^{l,g,p,q} \cdot \mathbf{Y}^{l+1}_{i+p, j+q}, \\
\mathbf{\tilde Y}^{l+1} &= \text{PixelShuffle}(\mathbf{\tilde Y}^{l+1, 1}, \mathbf{\tilde Y}^{l+1, 2}, \mathbf{\tilde Y}^{l+1, 3}, \mathbf{\tilde Y}^{l+1, 4}).
\end{aligned}
\end{equation}
}

As illustrated in Figure~\ref{fig:init_fusion}, the ALPF generator adaptively predicts spatially variant low-pass filters based on the feature content to smooth and enhance feature consistency. To provide deeper insights, visualized results are presented in Figure~\ref{fig:feat_sim_analysis}.
The findings depicted in Figure~\ref{fig:feat_sim_analysis}{\color{red}(a)} reveal that the commonly used bilinear upsampled feature in standard feature fusion exhibits significant intra-category inconsistency and boundary displacement. For instance, the car's interior shows low intra-category similarity, and the boundary appears blurred, indicating severe displacement.

{\color{revise}
In contrast, Figure~\ref{fig:feat_sim_analysis}{\color{red}(b)} demonstrates improved features characterized by enhanced interior consistency, which can be attributed to the introduction of the ALPF generator. This component effectively mitigates intra-category inconsistency, resulting in more cohesive features. Additionally, there is a noticeable improvement in boundary sharpness.

Quantitative analysis, as presented in Table~\ref{tab:sim_analysis}, corroborates these observations. 
Standard feature fusion techniques exhibit relatively low intra-category similarity, similarity margin, and similarity accuracy, thus increasing the risk of misclassification. 
However, the incorporation of the ALPF generator within the FreqFusion framework yields notable improvements. Specifically, there is a substantial increase in overall intra-category similarity (0.727$\rightarrow$0.799), similarity margin (0.245$\rightarrow$0.297), and similarity accuracy (0.918$\rightarrow$0.941).
In summary, the ALPF generator plays a pivotal role in enhancing feature consistency, thereby bolstering the effectiveness of the FreqFusion approach.
}

\begin{figure}[t!]
\centering
\scalebox{0.9918}{
\begin{tabular}{cccccc}
\includegraphics[width=0.918\linewidth]{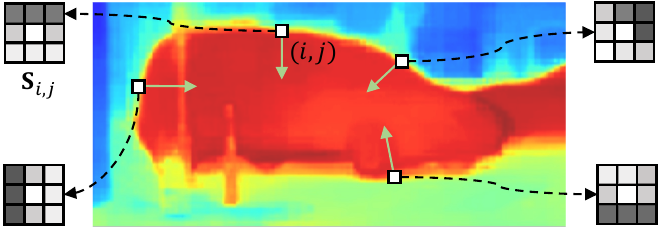} 
\end{tabular}
}
\caption{
\color{revise}
Illustration of how local similarity guides the offset prediction for resampling.
The brighter color indicates a higher intra-category similarity.
The $3\times 3$ gray grid indicates the cosine similarity $\mathbf{S}_{i, j}$ between the pixel at $(i, j)$ and its 8 local neighbors, including itself, for better visualization. 
A brighter color indicates a higher similarity.
Local similarity guides the offset generator to sample towards features with high intra-category similarity, thereby reducing the ambiguity in boundary or intra-category inconsistent areas.
}
\label{fig:sim_guide}
\end{figure}

\begin{figure}[t!]
\centering
\includegraphics[width=0.9818\linewidth]{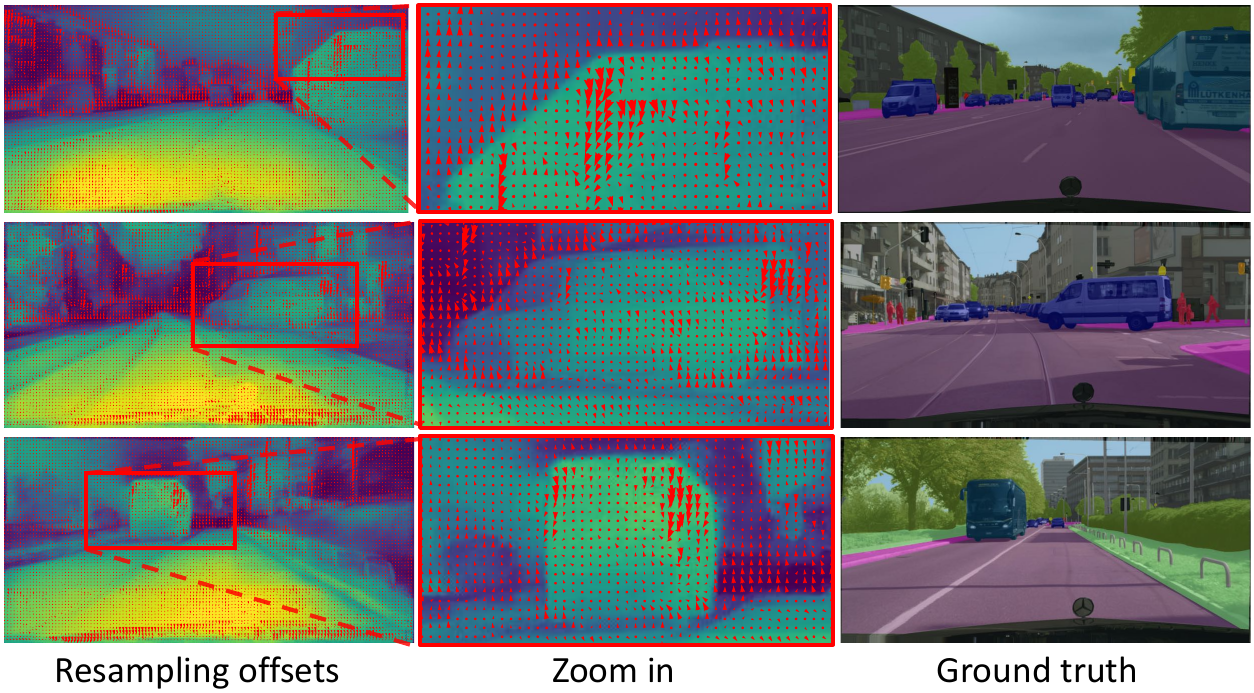} 
\caption{
Visualization for resampling offsets predicted by the offset generator.
The offsets at the inside boundaries of the buses and cars are toward the interior places where the features are more consistent and clear, while the offsets at the inside boundaries are towards the contrary direction, which makes the boundary clearer.
}
\label{fig:resampling_vis}
\end{figure}

{\color{revise}
\subsection{Offset Generator}
While the ALPF generator smooths features to enhance overall intra-category similarity, it may fall short in rectifying extensive areas of inconsistent features or refining thin and boundary areas.
Increasing the size of low-pass filters proves beneficial for addressing large inconsistent regions but can adversely impact thin and boundary areas. Conversely, reducing the filter size aids in preserving thin and boundary areas but may hinder the correction of extensive areas with inconsistent features.

To address this dilemma, we propose the offset generator. Motivated by the observation that neighboring features with low intra-category similarity often exhibit features with high intra-category similarity. 
The offset generator begins the process by computing local cosine similarity:
\begin{equation}
\begin{aligned}
\mathbf{S}^{l, p, q}_{i, j} = \frac{\sum_{c=1}^{C} \mathbf{Z}^{l}_{c, i, j} \cdot \mathbf{Z}^{l}_{c, i+p, j+q}}{\sqrt{\sum_{c=1}^{C}(\mathbf{Z}^{l}_{c,i, j})^2} \sqrt{\sum_{c=1}^{C}(\mathbf{Z}^{l}_{c, i+p, j+q})^2}},
\end{aligned}
\end{equation}
where $\mathbf{S}\in \mathbb{R}^{8\times H \times W}$ contain the cosine similarity between each pixel and its 8 neighbor pixels, which encourage the offset generator to sample towards features with high intra-category similarity, thereby reducing the ambiguity in boundary or intra-category inconsistent areas, as depicted in Figures~\ref{fig:sim_guide} and~\ref{fig:resampling_vis}.

Specifically, the offset generator takes the $\mathbf{Z}^{l}$ and $\mathbf{S}$ as input and predicts offsets. It consists of two $3\times 3$ convolutional layers to predict the offset direction and offset scale, represented as:
\begin{equation}
\begin{aligned}
\mathbf{O}^{l} &= \mathbf{D}^{l} \cdot \mathbf{A}^{l}, \\
\mathbf{D}^{l} &= \text{Conv}{3\times3}(\text{Concat}(\mathbf{Z}^{l}, \mathbf{S}^{l})), \\
\mathbf{A}^{l} &= \text{Sigmoid}(\text{Conv}{3\times3}(\text{Concat}(\mathbf{Z}^{l}, \mathbf{S}^{l}))),
\end{aligned}
\end{equation}
where $\mathbf{D}^{l}\in \mathbb{R}^{2G\times H \times W}$ represents the direction of offsets, $\mathbf{A}^{l}\in \mathbb{R}^{2G\times H \times W}$ aims to control the magnitude of offsets, and $\mathbf{O}^{l}\in \mathbb{R}^{2G\times H \times W}$ is the final predicted offsets for each pixel of the high-level feature.
$G$ denotes the number of offset groups; we strategically divide the feature into distinct groups, assigning unique spatial offsets for a more granular resampling.
This approach allows for resampling features with high intra-category similarity to replace features with low intra-category similarity. In this way, the offset generator can address large areas of inconsistent features and refine the boundary.

As shown in Figure~\ref{fig:resampling_vis}, at the inner boundaries of buses and cars, our offset generator strategically directs offsets toward interior locations where features exhibit higher consistency and clarity. Conversely, at the outer boundaries, we observe the offsets being strategically directed in the opposite direction, enriching the boundary regions with enhanced clarity. This intentional divergence in offset direction serves to accentuate the object boundaries.
Consequently, as shown in Figure~\ref{fig:feat_sim_analysis}{\color{red}(c)}, the offset generator contributes to achieving more consistent features and more accurate boundary delineations.
Quantitative analysis in Table~\ref{tab:sim_analysis} reveals that it enhances intra-category similarity (0.760$\rightarrow$0.799) and enhances similarity accuracy both overall (0.925$\rightarrow$0.941) and at the boundary (0.720$\rightarrow$0.728). This suggests that the offset generator provides benefits in addressing intra-category inconsistency and boundary displacement issues.
}

{\color{revise}
\subsection{Adaptive High-Pass Filter Generator}
\label{sec:AHPF}
Although the ALPF generator and offset generator effectively recover upsampled high-level features with high intra-class consistency and refined boundaries, the detailed boundary information present in lower-level features, lost during downsampling, cannot be fully restored in high-level features.

According to the Nyquist-Shannon Sampling Theorem \cite{shannon1949communication, nyquist1928}, frequencies higher than the Nyquist frequency, which is equivalent to half of the sampling rate, are permanently lost during downsampling. 
For example, when the high-level feature is downsampled by a factor of 2 compared to the low-level feature to be fused (\eg, using a 1$\times$1 convolution layer with a stride of 2 for downsampling, resulting in a sampling rate of $\frac{1}{2}$), frequencies above $\frac{1}{4}$ become aliased during the process.

To elaborate, we transform the feature map $\mathbf{X}\in \mathbb{R}^{C\times H\times W}$ into the frequency domain using the Discrete Fourier Transform (DFT), denoted as $\mathbf{X}_F = \mathcal{F}(\mathbf{X})$, expressed as:
\begin{equation}
\begin{aligned}
\mathbf{X}_F(u, v) = \frac{1}{HW}\sum_{h=0}^{H-1}\sum_{w=0}^{W-1}\mathbf{X}(h, w)e^{-2\pi j (uh + vw)},
\end{aligned}
\end{equation}
where $\mathbf{X}_F \in \mathbb{R}^{C\times H \times W}$ represents the output array of complex numbers from the DFT. $H$ and $W$ denote its height and width.
$h$, $w$ indicate the coordinates of feature map $\mathbf{X}$.
The normalized frequencies in the height and width dimensions are given by $\left|u\right|$ and $\left|v\right|$.
Consequently, the set of high frequencies larger than the Nyquist frequency $\mathcal{H}^{+} = \{(u, v) \mid \left|k \right| > \frac{{1}}{4} \text{ or } \left|l \right| > \frac{{1}}{4}\}$ is aliased and permanently lost in downsampled high-level features.

To address this limitation, we employ the AHPF generator to enhance detailed boundary information lost during downsampling. 
Specifically, the AHPF generator takes the initially fused $\mathbf{Z}^{l}$ as input and predicts spatial-variant high-pass filters. It consists of a $3\times 3$ convolutional layer followed by a softmax layer and a filter inversion operation, represented as:
\begin{equation}
\begin{aligned} 
\mathbf{\hat V}^{l} &= \text{Conv}_{3\times 3}(\mathbf{Z}^{l}), \\
\mathbf{\hat W}_{i, j}^{l,p,q} 
&=  \mathbf{E} - \text{Softmax}(\mathbf{\hat V}^{l}_{i, j}) \\
&= \mathbf{E}^{p,q} - \frac{\exp(\mathbf{\hat V}_{i, j}^{l,p,q})}{\sum_{p,q\in \Omega} \exp(\mathbf{\hat V}_{i, j}^{l,p,q})}, 
\end{aligned}
\end{equation}
where $\mathbf{\hat V}^{l}\in \mathbb{R}^{\hat K^2 \times H \times W}$ contains initial kernels at each location $(i, j)$.
$\hat K$ indicates the kernel size of high-pass filters.
To ensure the final generated kernels $\mathbf{\hat W}^{l}$ are high-pass, we follow~\cite{2021dynamic} to obtain low-pass kernels with kernel-wise softmax first and then invert the kernels by subtracting them from the identity kernel $\mathbf{E}$, whose weights are $[[0, 0, 0], [0, 1, 0], [0, 0, 0]]$ when $\hat K=3$.
After applying high-pass filters and adding residually, we obtain the enhanced results expressed as:
\begin{equation}
\begin{aligned}
\mathbf{\tilde X}^{l}_{i, j} = \mathbf{X}^{l}_{i, j} + \sum_{p,q\in \Omega} \mathbf{\hat W}_{i, j}^{l,p,q} \cdot \mathbf{X}^{l}_{i, j}.
\end{aligned}
\end{equation}
}

\begin{figure}[t!]
\color{revise}
\centering
\scalebox{0.9918}{
\begin{tabular}{cccccc}
\hspace{-2.918mm}
\includegraphics[width=0.9918\linewidth]{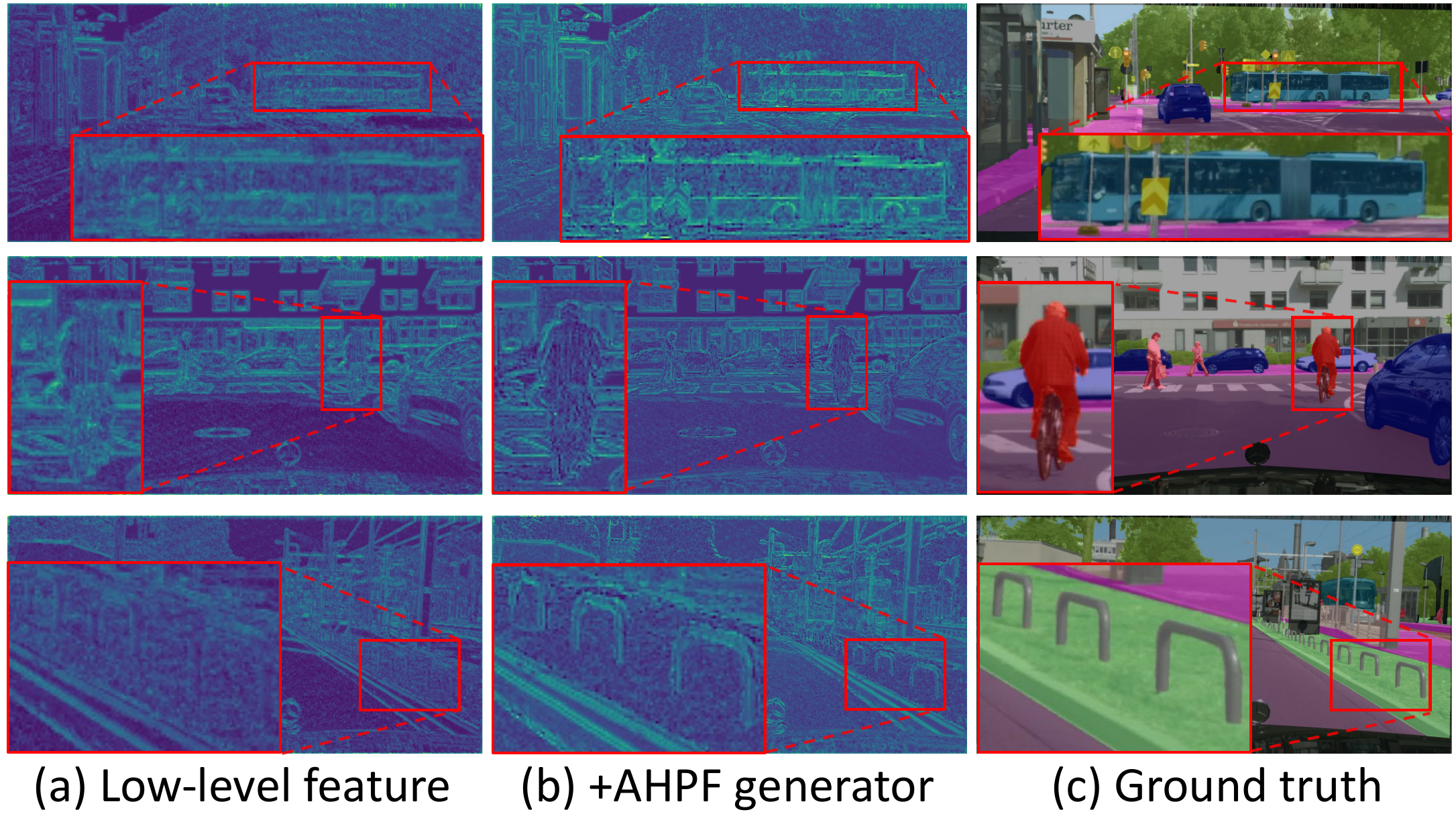} 
\end{tabular}
}
\caption{
Visualization of low-level features.
The AHPF generator significantly enhances boundaries by dynamically extracting high-frequency information from the low level, resulting in clearer object boundaries.
The red box indicates the zoom-in area.
}
\label{fig:feat_low}
\end{figure}

\begin{figure}[t!]
\centering
\scalebox{0.9918}{
\begin{tabular}{cccccc}
\hspace{-5mm} 
\includegraphics[width=0.9918\linewidth]{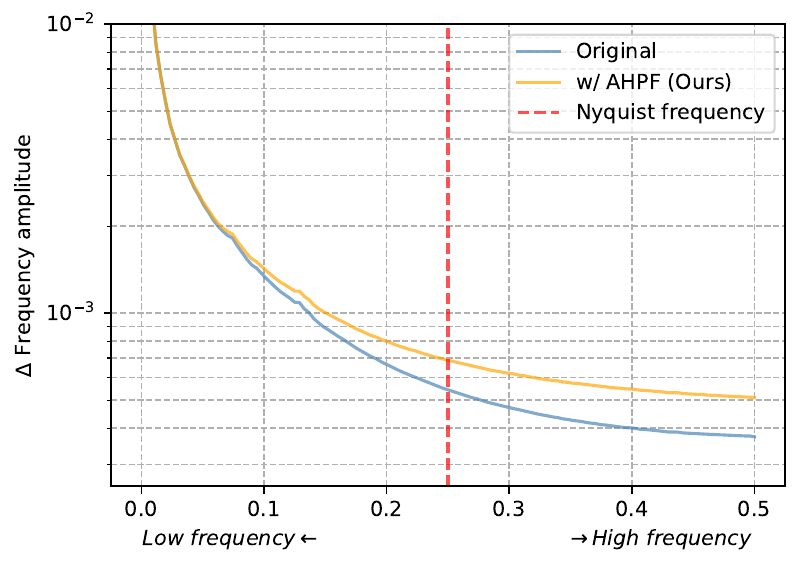} 
\end{tabular}
}
\caption{
Quantitative frequency analysis.
We present the results on a logarithmic scale for better visualization.
It indicates that the AHPF generator enhances high-frequency power, thereby improving the level of detail in the features and producing clearer boundaries.
}
\label{fig:feat_analysis}
\end{figure}

{\color{revise}
In Figure~\ref{fig:feat_low}, the effectiveness of the AHPF generator in enhancing detailed boundary information is evident. 
For instance, the original feature lacks clarity in delineating the outline of the bus and the details of a person's head. However, with the incorporation of the AHPF generator, these boundary details are substantially improved, resulting in a finer and more refined lower-level feature. 
The enhancement introduced by the AHPF generator highlights its capability to capture and preserve intricate details and boundaries, which are crucial for tasks requiring high-resolution and accurate feature representations. 
These visualizations corroborate and align with the quantitative frequency analysis presented in Figure~\ref{fig:feat_analysis}, which illustrates that the AHPF generator enhances high-frequency power above the Nyquist frequency.

Quantitative analysis in Table~\ref{tab:sim_analysis} demonstrates that it enhances the boundary similarity margin (0.228$\rightarrow$0.239) and boundary similarity accuracy (0.718$\rightarrow$0.728). 
This suggests that the AHPF generator provides benefits in addressing boundary displacement issues.
}

\section{Experimental Results}
\label{sec:experiments}
We first showcase the universality of the proposed FreqFusion across four typical dense prediction tasks, including semantic segmentation, object detection, instance segmentation, and panoptic segmentation. 
Following the setting~\cite{2023dysample, sapa} we set the kernel size in Deconvolution and Pixel Shuffle~\cite{2016pixelshuffle} as 3. 
For CARAFE~\cite{carafe}, we adhere to its default configuration. We utilize the `HIN' version of IndexNet~\cite{2019indexnet} and the `dynamic-cs-d†' version of A2U~\cite{2021a2u}. In the interest of stability across all dense prediction tasks, we opt for FADE~\cite{fade} without a gating mechanism and SAPA-B~\cite{sapa}.

\subsection{Semantic Segmentation}
Semantic segmentation necessitates the prediction of per-pixel class labels, ensuring that pixel groups belonging to the same object class are appropriately clustered. Typically, the decoder of a segmentation model employs a stage-by-stage upsampling and fusion architecture~\cite{2021segformer, 2018upernet}, highlighting the crucial role of feature fusion in this process.

Given the significance of feature fusion, FreqFusion is particularly well-suited to justify its behaviors in the context of semantic segmentation tasks. 
The inherent requirements of this task involve the precise clustering and separating of pixels for distinct object classes. 
This necessitates both low intra-category inconsistency and low boundary displacement, underscoring the importance of effective feature fusion mechanisms, such as those employed by FreqFusion.

\subsubsection{Experimental Settings}
\noindent{\bf Datasets.}
We evaluate our methods on several popular challenging datasets including Citysacpes~\cite{cityscapes2016}, ADE20K~\cite{ade20k}, and COCO-Stuff~\cite{2018cocostuff}.
Citysacpes~\cite{cityscapes2016} contains 19 semantic categories for semantic segmentation tasks and consists of 5,000 finely annotated images of $2048 \times 1024$ pixels, its training, validation, and test set have 2,975, 500, and 1,525 samples, respectively. 
We only use the training set for learning.
ADE20K~\cite{ade20k} is a challenging dataset that contains 150 semantic classes. 
It consists of 20,210, 2,000, and 3,352 images for the training, validation, and test sets. 
COCO-Stuff~\cite{2018cocostuff} is a challenging benchmark, which contains 172 semantic categories and 164k images in total.
\ie, 118k for training, 5k for validation, 20k for test-dev and 20k for the test-challenge.

\vspace{+1mm}
\noindent\textbf{Metrics.} 
In line with previous works such as Segformer~\cite{2021segformer}, Mask2Former~\cite{2022mask2former}, and SegNext~\cite{segnext}, we evaluate segmentation quality using the mean Intersection over Union (mIoU) for overall results and boundary Intersection over Union (bIoU)~\cite{2021boundaryiou} for boundary delineation. 
Additionally, we present results regarding the number of GFLOPs and parameters to facilitate a comprehensive comparison of computational and storage costs.

\vspace{+1mm}
\noindent\textbf{Implementation Details.} 
When applying the proposed method to existing methods (SegFormer~\cite{2021segformer}, Mask2Former~\cite{2022mask2former}, SegNeXt~\cite{segnext}, etc.), we adopt their original training setting.
\eg, for SegFormer~\cite{2021segformer} and SegNeXt~\cite{segnext}, 
we use common data augmentation including random horizontal flipping, random resizing from 0.5 to 2, and random cropping (1024$\times$1024 on Cityscapes, 512$\times$512 on ADE20K and COCO-Stuff). 
The batch size is set to 8 for the Cityscapes dataset and 16 for all the other datasets. 
AdamW~\cite{2017adamw} is applied to train our models. 
The initial learning rate is 0.00006 and the poly-learning rate decay policy~\cite{deeplabv3plus} is employed. 
We train our model 160K iterations for ADE20K, and Cityscapes, and 80K iterations for COCO-Stuff.
For SegFormer~\cite{2021segformer} and Mask2Former~\cite{2022mask2former}, which fuse 4$\times$, 8$\times$, 16$\times$, and 32$\times$ downsampled features, we utilize 3 FreqFusion modules. 
For SegNeXt~\cite{segnext}, which fuses 8$\times$, 16$\times$, and 32$\times$ downsampled features, we employ 2 FreqFusion modules.
\begin{figure}[t!]
\color{revise}
\centering
\scalebox{0.99999998}{
\begin{tabular}{cccccc}
\hspace{-2.918mm}
\includegraphics[width=0.98\linewidth]{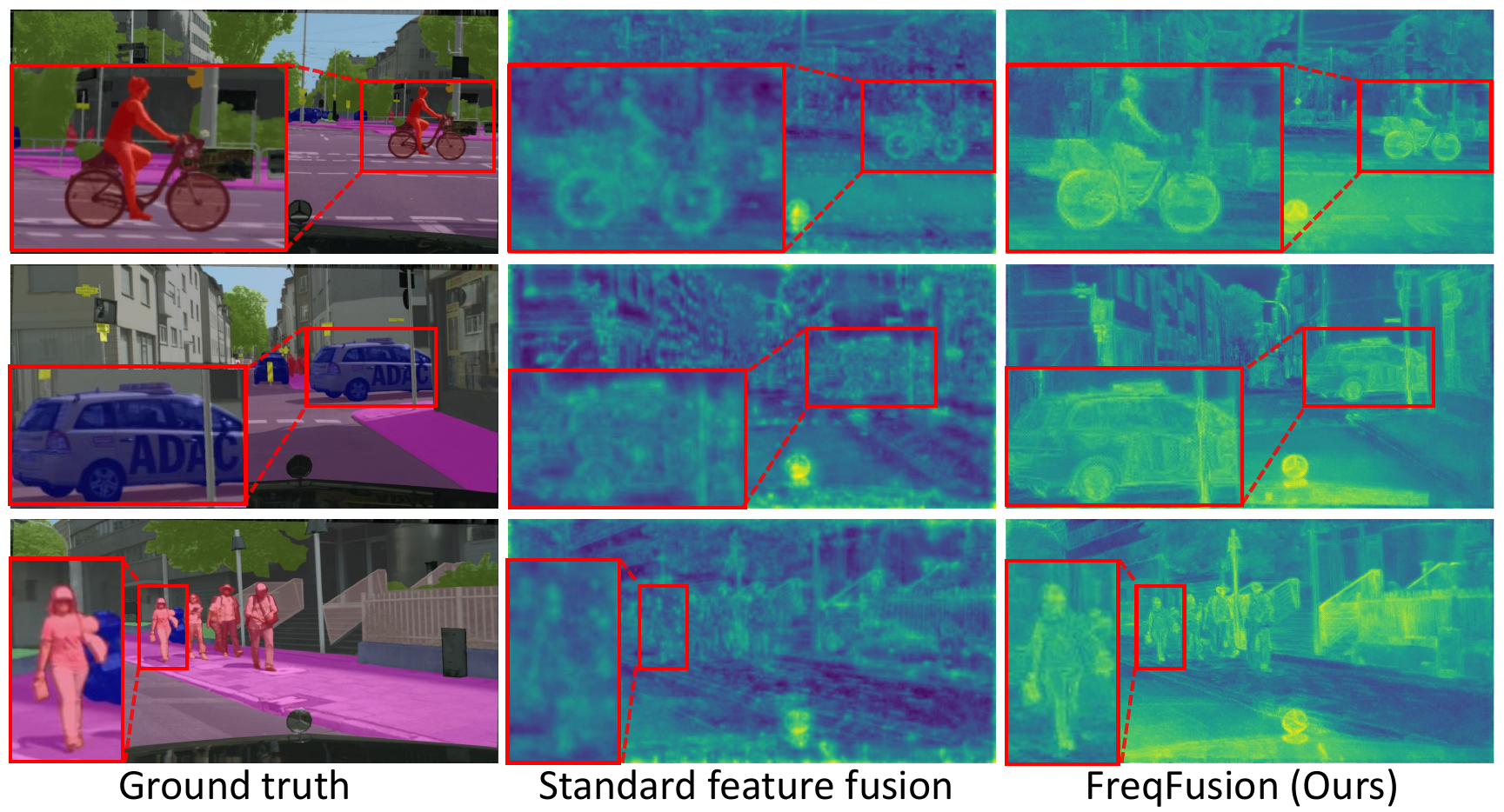} 
\end{tabular}
}
\caption{
Visualization for feature fusion.
In comparison with standard feature fusion, the fused features obtained by FreqFusion exhibit more consistent features with sharper boundaries.
The red boxes indicate the zoom-in areas.
}
\label{fig:feature_fusion}
\end{figure}

\begin{figure*}[t!]
\color{revise}
\centering
\includegraphics[width=0.9918\linewidth]{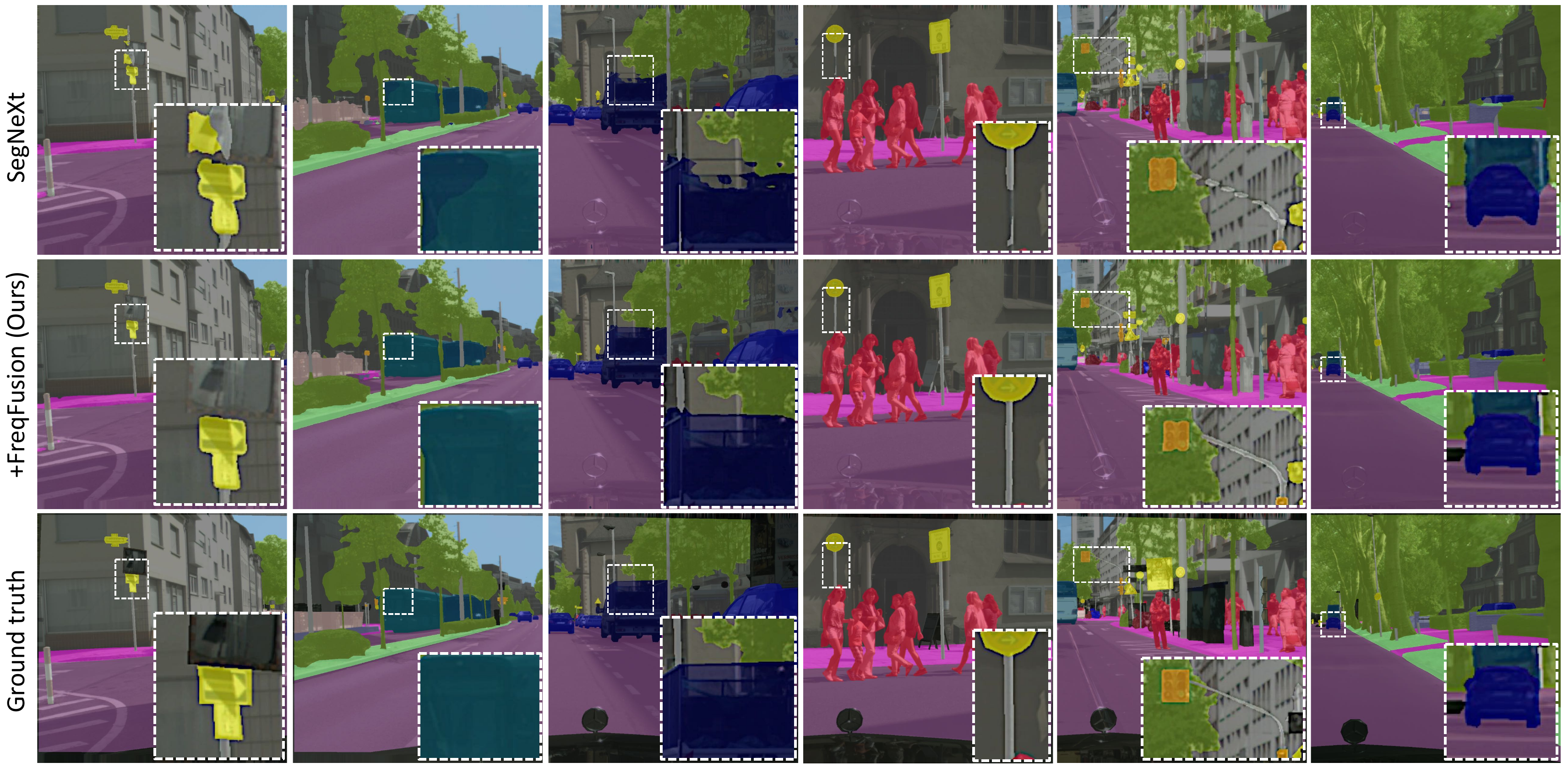}    
\caption{
Visualization on the Cityscapes~\cite{cityscapes2016} validation set. 
Compared with the vanilla model SegNeXt~\cite{segnext} (row one), the proposed FreqFusion (row two) considerably improves the segmentation accuracy and consistency.
}
\label{fig:city_vis}
\end{figure*}
\begin{figure*}[t!]
\color{revise}
\centering
\includegraphics[width=0.9918\linewidth]{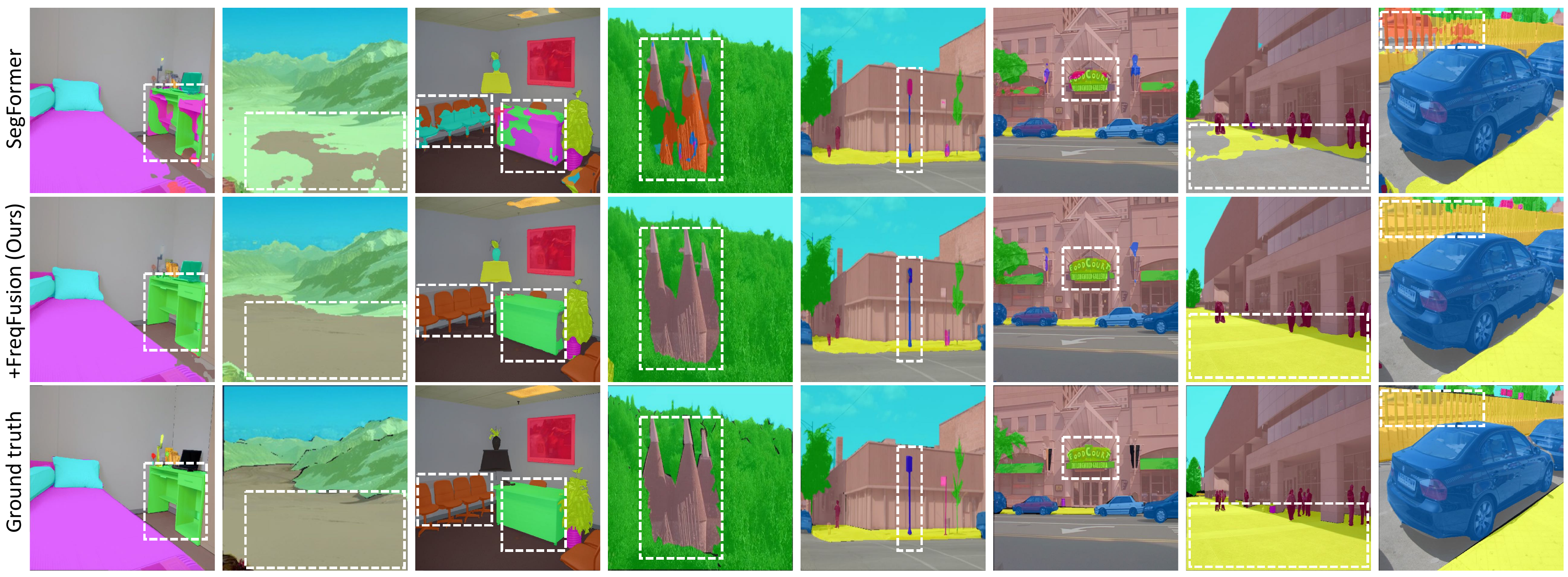}    
\caption{
Visualization on the ADE20K~\cite{ade20k} validation set. 
Compared with the vanilla model SegFormer~\cite{2021segformer} (row one), the proposed FreqFusion (row two) considerably improves the segmentation accuracy and consistency.
}
\label{fig:ade20k_vis}
\end{figure*}

\begin{figure*}[t!]
\color{revise}
\centering
\includegraphics[width=0.9918\linewidth]{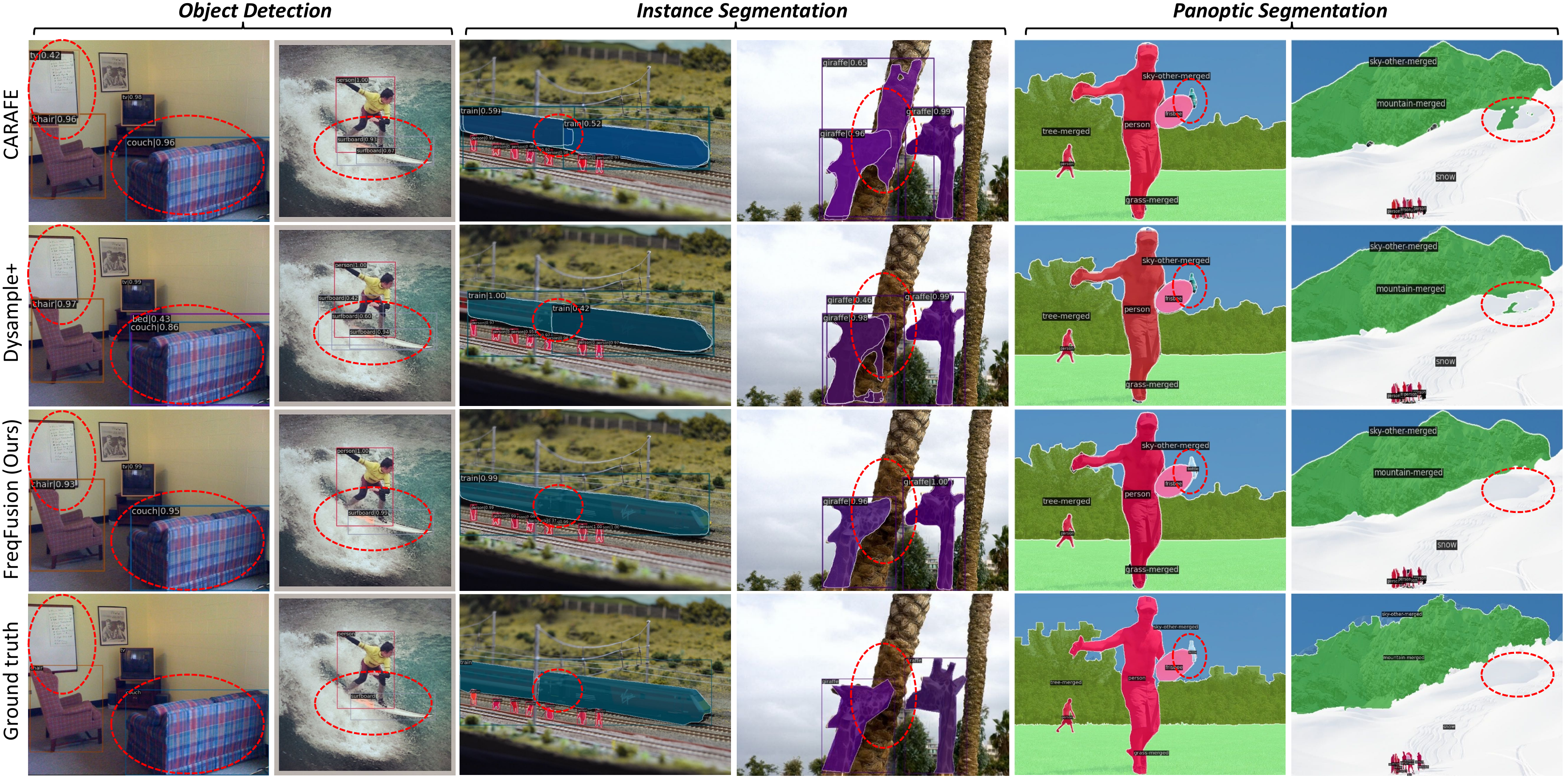}    
\caption{
Visualization on the COCO~\cite{2014microsoft} validation set.
We select the top two best-performing methods, CARAFE~\cite{carafe} (row one) and Dysample~\cite{2023dysample} (row two), excluding our proposed FreqFusion.
In comparison, our FreqFusion (row three) demonstrates superior prediction accuracy and consistency.
}
\label{fig:coco}
\end{figure*}

\begin{table}[tb!]
\centering
\caption{
Comparison with recent state-of-the-art methods on the ADE20K~\cite{ade20k} validation set.
We present mIoU and bIoU results to assess the intra-category consistency and boundary displacement of the final predictions.
}
\scalebox{0.918}{
\begin{tabular}{l|l|l|cccccc}
\toprule[1.28pt]
Method & Params (M)& FLOPs (G) & mIoU & bIoU \\
\midrule
Segformer-B1~{\color{gray}\tiny [NeurIPS2021]}~\cite{2021segformer} &13.74 &15.91 & 41.7 & 27.8   \\
\midrule
+ Deconv{\color{gray}\tiny [CVPR2010]}~\cite{2010deconv} &+3.5 &+34.4 & 40.7 & 25.9 \\
+ PixelShuffle{\color{gray}\tiny [CVPR2016]}~\cite{2016pixelshuffle} &+14.2 &+34.4 & 41.5 & 26.6\\
+ CARAFE{\color{gray}\tiny [ICCV2019]}~\cite{carafe} &+0.44 &+1.45 & 42.8 & 29.8 \\
+ IndexNet{\color{gray}\tiny [TPAMI2020]}~\cite{2020indexnet}  &+12.6 & +30.65 & 41.5 & 28.3\\
+ A2U{\color{gray}\tiny [CVPR2021]}~\cite{2021a2u} &+0.12  &+0.41 & 41.5 & 27.3 \\ 
+ FADE{\color{gray}\tiny [ECCV2022]}~\cite{fade} &+0.29 &+2.65 & 43.1 & 31.7\\
+ SAPA-B{\color{gray}\tiny [NeurIPS2022]}~\cite{sapa} &+0.1 &+1.0 & 43.2 & 31.0 \\
+ Dysample-S+{\color{gray}\tiny [ICCV2023]}~\cite{2023dysample} &+0.01 &+0.3 & 43.6 & 29.9\\
+ Dysample+{\color{gray}\tiny [ICCV2023]}~\cite{2023dysample} &+0.1 &+0.4 & 43.3 & 29.2\\
\midrule
\rowcolor{mygray!58}
+ FreqFusion (Ours) & +0.34 & +2.35 & \bf 44.5 &\bf 32.8 \\
\bottomrule[1.28pt]
\end{tabular}
}
\label{tab:featupsampling}
\end{table}

\begin{table}[tb!]
\centering
\caption{
Comparison with recent state-of-the-art dynamic sampling-based segmentation methods on Cityscapes~\cite{cityscapes2016} validation set.
}
\scalebox{0.95188}{
\begin{tabular}{l|l|ccccccc}
\toprule[1.28pt]
Method & Backbone & mIoU  \\
\midrule
AlignSeg~{\color{gray}\tiny [TPAMI2021]}~\cite{2021alignseg} & ResNet-50 & 78.5 \\
IFA~{\color{gray}\tiny [ECCV2022]}~\cite{2022IFA} & ResNet-50  & 78.0  \\
SFNet~{\color{gray}\tiny [IJCV2023]}~\cite{2023sfnet} & ResNet-50  & 79.2   \\
FaPN~{\color{gray}\tiny [ICCV2021]}~\cite{2021fapn}  & ResNet-50 & 80.0 \\
Mask2Former{\color{gray}\tiny [CVPR2022]}~\cite{2022mask2former} & ResNet-50 & 79.4 \\
FaPN~{\color{gray}\tiny [ICCV2021]}~\cite{2021fapn}  & ResNet-101 & 80.1 \\
Mask2Former{\color{gray}\tiny [CVPR2022]}~\cite{2022mask2former}& ResNet-101 & 80.1 \\
\midrule
\rowcolor{mygray!58}
Mask2Former~\cite{2022mask2former} + FreqFusion (Ours)& ResNet-50 & \bf 80.5 (+1.4)\\
\bottomrule[1.28pt]
\end{tabular}
}
\label{tab:feature_align}
\end{table}

\begin{table}[tb!]
\centering
\caption{
The combination of the proposed FreqFusion and various state-of-the-art model structures.
Results are reported on Cityscapes~\cite{cityscapes2016} validation set with single scale inference.
The FLOPs results are calculated with an image size of $1024\times 2048$.
}
\scalebox{0.7218}{
\begin{tabular}{l|l|cc|cc|cccc}
\toprule[1.28pt]
\multirow{2}{*}{Method} & \multirow{2}{*}{Backbone} & \multicolumn{2}{c|}{Parameters (M)} & \multicolumn{2}{c|}{FLOPs (G)} & \multicolumn{2}{c}{mIoU}\\
\cmidrule{3-8}
& & Vanilla & Ours  & Vanilla & Ours & Vanilla & Ours \\
\midrule
UPerNet~\cite{2018upernet} & ResNet-50 &28.8 &29.0 &300.9 & 315.8 & 78.8 &\bf 79.8 (+1.0) \\
SegFormer\cite{2021segformer} &MiT-B1 &13.7 &14.1 &243.7 & 271.8 & 78.5 & \bf 80.1 (+1.6) \\
SegNeXt~\cite{segnext} &MSCAN-T &4.3 &4.4 &50.5 &52.4 & 79.8 &\bf 80.8 (+1.0) \\
\bottomrule[1.28pt]
\end{tabular}
}
\label{tab:combineSOTA}
\end{table}

\begin{table}[tb!]
\color{revise}
\centering
\caption{
Semantic segmentation results with recent state-of-the-art large model Mask2Former~\cite{2022mask2former} on ADE20K. The best performance is in boldface.
}
\scalebox{0.9188}{
\begin{tabular}{l|l|ccccccc}
\toprule[1.28pt]
Mask2Former{\color{gray}\tiny [CVPR2022]}~\cite{2022mask2former} & Backbone & mIoU \\
\midrule
Bilinear & Swin-B$^\dagger$ & 53.9 \\
\rowcolor{mygray!58}
FreqFusion (Ours) & Swin-B$^\dagger$ &\bf 55.3 (+1.4) \\
\midrule
Bilinear & Swin-L$^\dagger$ & 56.1  \\
\rowcolor{mygray!58}
FreqFusion (Ours) & Swin-L$^\dagger$ &\bf 56.8 (+0.7)\\
\bottomrule[1.28pt]
\end{tabular}
}
\label{tab:mask2former}
\end{table}

\begin{table}[tb!]
\centering
\caption{
Results reported on various challenging datasets with SegNeXt~\cite{segnext}, including Cityscapes\cite{cityscapes2016}, ADE20K\cite{ade20k}, and COCO-stuff~\cite{2018cocostuff}.
}
\scalebox{0.7518}{
\begin{tabular}{l|l|l|c|c|cccc}
\toprule[1.28pt]
\multirow{2}{*}{Method} & \multirow{2}{*}{Params}  & \multirow{2}{*}{FLOPs} & Cityscapes & ADE20K & COCO-stuff \\
& & &mIoU &mIoU &mIoU  \\
\midrule
SegNeXt-T{\color{gray}\tiny [NeurIPS2022]}~\cite{segnext} & 4.26M & 6.59G &79.8 & 41.1 & 38.7  \\
\rowcolor{mygray!58}
+ FreqFusion (Ours) & +0.18M & +0.44G &\bf 80.8 (+1.0) & \bf 43.5 (+2.4) &\bf 40.7 (+2.0)  \\
\bottomrule[1.28pt]
\end{tabular}
}
\label{tab:various_dataset}
\end{table}

\begin{table}[t!]
\centering
\color{revise}
\caption{
Comparison with recent state-of-the-art methods on the ADE20K~\cite{ade20k} validation set.
}
\scalebox{0.918}{
\begin{tabular}{l|l|l|cccccc}
\toprule[1.28pt]
Method & Params (M)& FLOPs (G) & mIoU  \\
\midrule
SegNeXt-T~{\color{gray}\tiny [NeurIPS2022]}~\cite{segnext} &4.26 &6.59 & 41.1   \\
\midrule
+ Dysample+{\color{gray}\tiny [ICCV2023]}~\cite{2023dysample} &+0.04 &+0.03 & 42.2\\
\midrule
\rowcolor{mygray!58}
+ FreqFusion (Ours) & +0.18 & +0.44 & \bf 43.5 & \\
\bottomrule[1.28pt]
\end{tabular}
}
\label{tab:segnext_comp}
\end{table}

\begin{table}[t!]
\centering
\color{revise}
\caption{
Inference speed evaluation.
We apply FreqFusion to efficient segmentation model SegNeXt-T~\cite{segnext}.
The frame per second (FPS) results are tested with an image size of $1024\times 2048$ on a single RTX 3090. 
FreqFusion shows a minor impact on the FPS.
}
\scalebox{0.7918}{
\begin{tabular}{l|c|c|cccccc}
\toprule[1.28pt]
 Model & SegNeXt-T~\cite{segnext} & +Dysample~\cite{2023dysample} & +FreqFusion (Ours) \\
\midrule
FPS &26.5  & 25.9 & 23.0 \\
\bottomrule[1.28pt]
\end{tabular}
}
\label{tab:fps}
\end{table}

\subsubsection{Semantic Segmentation Results}
\vspace{+1mm}
\noindent{\bf Comparison with previous state-of-the-art methods.}
As demonstrated in Table~\ref{tab:featupsampling}, employing the widely-used SegFormer-B1~\cite{2021segformer} as the segmentation model, FreqFusion achieves a 2.8 mIoU improvement on ADE20K, surpassing all previous state-of-the-art competitors, including CARAFE, IndexNet, A2U, FADE, SAPA-B, Dysample-S+, and Dysample+.
Notably, FreqFusion outperforms the second-place Dysample-S+ by a large margin of 1.2 mIoU.

\noindent{\bf Combination with state-of-the-art methods.}
As evidenced in Table~\ref{tab:feature_align}, when employing Mask2Former~\cite{2022mask2former} as the segmentation model, FreqFusion achieves a notable 1.4 mIoU improvement on the Cityscapes dataset, outperforming its competitors, including AlignSeg, IFA, SFNet, and the original Mask2Former, which adopt an offset-based mechanism. Remarkably, FreqFusion demonstrates superior performance even when using ResNet-50 as the backbone, surpassing Mask2Former with a heavier ResNet-101 by a margin of 0.4 mIoU.

\vspace{+1mm}
\noindent{\bf Combination with various model structures.}
In Table~\ref{tab:combineSOTA}, we apply FreqFusion to various state-of-the-art methods from recent CNNs (\eg, SegNeXt~\cite{segnext}) to Transformers (\eg, SegFormer~\cite{2021segformer}).
The UPerNet~\cite{2018upernet} adopt FPN~\cite{2017feature} structure, while SegFormer~\cite{2021segformer} and SegNeXt~\cite{segnext} use concatenation for feature fusion.
Though their structures show a large difference, our FreqFusion can consistently improve their performance with very minor extra parameters and computation.
This shows that FreqFusion generalizes well to various modern model structures. 

{\color{revise}
\vspace{+1mm}
\noindent{\bf Combination with large models.}
When applied to MaskFormer with a large backbone, FreqFusion demonstrates substantial improvements in the mIoU metric. Specifically, the mIoU increases from 53.9 to 55.3 (+1.4) when employing Swin-B and from 56.1 to 56.8 (+0.7) with Swin-L, as illustrated in Table~\ref{tab:mask2former}. This observation underscores the effectiveness of FreqFusion even when integrated with recent state-of-the-art heavy segmentation models. It implies that intra-category inconsistency and boundary displacement are widely prevalent challenges in these advanced models.
}

\vspace{+1mm}
\noindent{\bf Experiments with various challenging datasets.}
Utilizing SegNeXt~\cite{segnext} as the segmentation model, we conducted experiments on diverse and challenging datasets. The results, as presented in Table~\ref{tab:various_dataset}, consistently demonstrate that the proposed FreqFusion leads to improvements across multiple datasets. Specifically, FreqFusion improves SegNeXt~\cite{segnext} by 1.0, 2.4, and 2.0 on Cityscapes, ADE20K, and COCO-stuff, respectively.

{\color{revise}
\vspace{+1mm}
\noindent{\bf FPS results.}
Here, we provide additional FPS (frames per second) results for further efficiency analysis and comparison. 
As shown in Tables~\ref{tab:segnext_comp} and~\ref{tab:fps}, when combining our method with the state-of-the-art efficient segmentation model SegNeXt, FreqFusion introduces more parameters and GFLOPs than the recent state-of-the-art method Dysample~\cite{2023dysample}. However, FreqFusion demonstrates a much higher performance improvement (+2.4 mIoU \emph{vs.} +1.1 mIoU) than Dysample. Moreover, FreqFusion achieves an FPS of 23.0, which is very close to the fastest recent Dysample~\cite{2023dysample} with an FPS of 25.9. This indicates that while the proposed method is slightly slower, it achieves much higher accuracy, demonstrating the satisfactory efficiency of our method.
}

\vspace{+1mm}
\noindent{\bf Visual results.}
As shown in Figure~\ref{fig:feature_fusion}, we visualize the features. In comparison with standard feature fusion, the fused features obtained by FreqFusion exhibit more consistent features with sharper boundaries.
Additionally, in Figures~\ref{fig:city_vis} and~\ref{fig:ade20k_vis}, we present additional visualizations of segmentation results on Cityscapes~\cite{cityscapes2016} and ADE20K~\cite{ade20k}. Compared with the baseline model SegNeXt~\cite{segnext} and SegFormer~\cite{2021segformer}, which adopt standard feature fusion, the proposed FreqFusion significantly improves segmentation accuracy and consistency. These results validate the effectiveness of FreqFusion.

\begin{table}[tb!]
\centering
\caption{
Object detection results of Faster R-CNN with ResNet-50 on MS-COCO. Best performance is in boldface.
}
\scalebox{0.6918}{
\begin{tabular}{lcccccccccc}
\toprule[1.28pt]
Method & Backbone &Param. & $AP^{}$ & $AP_{50}^{}$ & $AP_{75}^{}$ & $AP_{S}^{}$ & $AP_{M}^{}$ & $AP_{L}^{}$ \\
\midrule
Nearest & R50 & 46.8M & 37.5 & 58.2 & 40.8 & 21.3 & 41.1 & 48.9 \\
Deconv & R50 & +2.4M & 37.3 & 57.8 & 40.3 & 21.3 & 41.1 & 48.0 \\
PixelShuffle{\color{gray}\tiny [CVPR2016]}~\cite{2016pixelshuffle} & R50 & +9.4M & 37.5 & 58.5 & 40.4 & 21.5 & 41.5 & 48.3 \\
CARAFE{\color{gray}\tiny [ICCV2019]}~\cite{carafe} & R50 & +0.3M & 38.6 & 59.9 & 42.2 & 23.3 & 42.2 & 49.7 \\
IndexNet{\color{gray}\tiny [TPAMI2020]}~\cite{2020indexnet} & R50 & +8.4M & 37.6 & 58.4 & 40.9 & 21.5 & 41.3 & 49.2 \\
A2U{\color{gray}\tiny [CVPR2021]}~\cite{2021a2u} & R50 & +38.9K & 37.3 & 58.7 & 40.0 & 21.7 & 41.1 & 48.5 \\
FADE{\color{gray}\tiny [ECCV2022]}~\cite{fade} & R50 & +0.2M & 38.5 & 59.6 & 41.8 & 23.1 & 42.2 & 49.3 \\
SAPA-B{\color{gray}\tiny [NeurIPS2022]}~\cite{sapa} & R50 & +0.1M & 37.8 & 59.2 & 40.6 & 22.4 & 41.4 & 49.1 \\
DySample-S+{\color{gray}\tiny [ICCV2023]}~\cite{2023dysample} & R50 & +8.2K & 38.6 & 59.8 & 42.1 & 22.5 & 42.1 & 50.0 \\
DySample+{\color{gray}\tiny [ICCV2023]}~\cite{2023dysample} & R50 & +65.5K & 38.7 & 60.0 & 42.2 & 22.5 & 42.4 & 50.2 \\
\rowcolor{mygray!58}
FreqFusion (Ours) & R50 & +0.3M &\bf 39.4 &\bf 60.9 &\bf 42.7 &\bf 23.0 &\bf 43.3 &\bf 50.9 \\
\midrule
Nearest & R101 & 65.8M & 39.4 & 60.1 & 43.1 & 22.4 & 43.7 & 51.1 \\
DySample+{\color{gray}\tiny [ICCV2023]}~\cite{2023dysample} & R101 & +65.5K & 40.5 & 61.6 & 43.8 & 24.2 & 44.5 & 52.3 \\
\rowcolor{mygray!58}
FreqFusion (Ours)& R101 & +0.3M &\bf 41.0 &\bf 62.2 &\bf 44.9 &\bf 24.7 &\bf 45.0 &\bf 53.4 \\
\bottomrule[1.28pt]
\end{tabular}
}
\label{tab:FasterRCNN}
\end{table}

\begin{table}[tb!]
\centering
\caption{
Instance segmentation results of Mask R-CNN with ResNet50 on MS-COCO. 
The parameter increment is identical as in Faster R-CNN. 
The upper table shows box AP results for detection, and the bottom table shows mask AP results for instance segmentation.
The best performance is in boldface.
}
\scalebox{0.7518}{
\begin{tabular}{lcccccccccc}
\toprule[1.28pt]
Method & Backbone & $AP^{}$ & $AP_{50}^{}$ & $AP_{75}^{}$ & $AP_{S}^{}$ & $AP_{M}^{}$ & $AP_{L}^{}$ \\
\midrule
Nearest & R50 & 38.3 & 58.7 & 42 & 21.9 & 41.8 & 50.2 \\
Deconv & R50 & 37.9 & 58.5 & 41.0 & 22.0 & 41.6 & 49.0 \\
PixelShuffle{\color{gray}\tiny [CVPR2016]}~\cite{2016pixelshuffle} & R50 & 38.5 & 59.4 & 41.9 & 22.0 & 42.3 & 49.8 \\
CARAFE{\color{gray}\tiny [ICCV2019]}~\cite{carafe} & R50 & 39.2 & 60.0 & 43.0 & 23.0 & 42.8 & 50.8 \\
IndexNet{\color{gray}\tiny [TPAMI2020]}~\cite{2020indexnet} & R50 & 38.4 & 59.2 & 41.7 & 22.1 & 41.7 & 50.3 \\
A2U{\color{gray}\tiny [CVPR2021]}~\cite{2021a2u} & R50 & 38.2 & 59.2 & 41.4 & 22.3 & 41.7 & 49.6 \\
FADE{\color{gray}\tiny [ECCV2022]}~\cite{fade} & R50 & 39.1 & 60.3 & 42.4 & 23.6 & 42.3 & 51.0 \\
SAPA-B{\color{gray}\tiny [NeurIPS2022]}~\cite{sapa} & R50 & 38.7 & 59.7 & 42.2 & 23.1 & 41.8 & 49.9 \\
DySample-S+{\color{gray}\tiny [ICCV2023]}~\cite{2023dysample} & R50 & 39.3 & 60.3 & 42.8 & 23.2 & 42.7 & 50.8 \\
DySample+{\color{gray}\tiny [ICCV2023]}~\cite{2023dysample} & R50 & 39.6 & 60.4 & 43.5 & 23.4 & 42.9 &\bf 51.7 \\
\rowcolor{mygray!58}
FreqFusion (Ours) & R50 &\bf 40.0 &\bf 61.2 &\bf 43.5 &\bf 24.3 &\bf 43.9 & 51.5 \\
\midrule
Nearest & R101 & 40.0 & 60.4 & 43.7 & 22.8 & 43.7 & 52.0 \\
DySample+{\color{gray}\tiny [ICCV2023]}~\cite{2023dysample} & R101 & 41.0 & 61.9 & 44.9 & 24.3 & 45.0 & 53.5 \\
\rowcolor{mygray!58}
FreqFusion (Ours) & R101 &\bf 41.6 &\bf 62.4 &\bf 45.7 &\bf 25.0 &\bf 45.6 &\bf 54.2 \\
\midrule
\multicolumn{8}{c}{\it Instance Segmentation Mask Results} \\
\midrule
Nearest & R50 & 34.7 & 55.8 & 37.2 & 16.1 & 37.3 & 50.8 \\
Deconv & R50 & 34.5 & 55.5 & 36.8 & 16.4 & 37.0 & 49.5 \\
PixelShuffle{\color{gray}\tiny [CVPR2016]}~\cite{2016pixelshuffle} & R50 & 34.8 & 56.0 & 37.3 & 16.3 & 37.5 & 50.4 \\
CARAFE{\color{gray}\tiny [ICCV2019]}~\cite{carafe} & R50 & 35.4 & 56.7 & 37.6 & 16.9 & 38.1 & 51.3 \\
IndexNet{\color{gray}\tiny [TPAMI2020]}~\cite{2020indexnet} & R50 & 34.7 & 55.9 & 37.1 & 16.0 & 37.0 & 51.1 \\
A2U{\color{gray}\tiny [CVPR2021]}~\cite{2021a2u} & R50 & 34.6 & 56.0 & 36.8 & 16.1 & 37.4 & 50.3 \\
FADE{\color{gray}\tiny [ECCV2022]}~\cite{fade} & R50 & 35.1 & 56.7 & 37.2 & 16.7 & 37.5 & 51.4 \\
SAPA-B{\color{gray}\tiny [NeurIPS2022]}~\cite{sapa} & R50 & 35.1 & 56.5 & 37.4 & 16.7 & 37.6 & 50.6 \\
DySample-S+{\color{gray}\tiny [ICCV2023]}~\cite{2023dysample} & R50 & 35.5 & 56.8 & 37.8 & 17.0 & 37.9 & 51.9 \\
DySample+{\color{gray}\tiny [ICCV2023]}~\cite{2023dysample} & R50 & 35.7 & 57.3 &\bf 38.2 & 17.3 & 38.2 & 51.8 \\
\rowcolor{mygray!58}
FreqFusion (Ours) & R50 &\bf 36.0 &\bf 57.9 & 38.1 &\bf 17.9 &\bf 39.0 &\bf52.3 \\
\midrule
Nearest & R101 & 36.0 & 57.6 & 38.5 & 16.5 & 39.3 & 52.2 \\
DySample+{\color{gray}\tiny [ICCV2023]}~\cite{2023dysample} & R101 & 36.8 & 58.7 & 39.5 & 17.5 & 40.0 & 53.8 \\
\rowcolor{mygray!58}
FreqFusion (Ours) & R101 & \bf 37.4 &\bf 59.6 &\bf 39.9 &\bf 18.4 &\bf 40.6 &\bf 54.3 \\
\bottomrule[1.28pt]
\end{tabular}
}
\label{tab:MaskRCNN}
\end{table}

\begin{table}[tb!]
\centering
\caption{
Panoptic segmentation results with Panoptic FPN on MS-COCO. 
The best performance is in boldface.
}
\scalebox{0.7518}{
\begin{tabular}{lcccccccccc}
\toprule[1.28pt]
Method & Backbone & Params & $PQ$ & $PQ^{th}$ & $PQ^{st}$ & $SQ$ & $RQ$ \\
\midrule
Nearest & R50 & 46.0M & 40.2 & 47.8 & 28.9 & 77.8 & 49.3 \\
Deconv & R50 & +1.8M & 39.6 & 47.0 & 28.4 & 77.1 & 48.5 \\
PixelShuffle{\color{gray}\tiny [CVPR2016]}~\cite{2016pixelshuffle} & R50 & +7.1M & 40.0 & 47.4 & 28.8 & 77.1 & 49.1 \\
CARAFE{\color{gray}\tiny [ICCV2019]}~\cite{carafe} & R50 & +0.2M & 40.8 & 47.7 & 30.4 & 78.2 & 50.0 \\
IndexNet{\color{gray}\tiny [TPAMI2020]}~\cite{2020indexnet} & R50 & +6.3M & 40.2 & 47.6 & 28.9 & 77.1 & 49.3 \\
A2U{\color{gray}\tiny [CVPR2021]}~\cite{2021a2u} & R50 & +29.2K & 40.1 & 47.6 & 28.7 & 77.3 & 48.0 \\
FADE{\color{gray}\tiny [ECCV2022]}~\cite{fade} & R50 & +0.1M & 40.9 & 48.0 & 30.3 & 78.1 & 50.1 \\
SAPA-B{\color{gray}\tiny [NeurIPS2022]}~\cite{sapa}& R50 & +0.1M & 40.6 & 47.7 & 29.8 & 78.0 & 49.6 \\
DySample-S+{\color{gray}\tiny [ICCV2023]}~\cite{2023dysample} & R50 & +6.2K & 41.1 & 48.1 & 30.5 & 78.2 & 50.2 \\
DySample+{\color{gray}\tiny [ICCV2023]}~\cite{2023dysample} & R50 & +49.2K & 41.5 & 48.5 & 30.8 & 78.3 & 50.7 \\
\rowcolor{mygray!58}
FreqFusion (Ours) & R50 & +0.3M &\bf 42.7 &\bf 49.3 &\bf 32.7 &\bf 79.0 &\bf 51.9 \\
\midrule
Nearest & R101 & 65.0M & 42.2 & 50.1 & 30.3 & 78.3 & 51.4 \\
DySample+{\color{gray}\tiny [ICCV2023]}~\cite{2023dysample} & R101 & +49.2K & 43.0 & 50.2 & 32.1 & 78.6 & 52.4 \\
\rowcolor{mygray!58}
FreqFusion (Ours) & R101 & +0.3M & \bf 44.0 & \bf 50.8 & \bf 33.7 & \bf 79.4 & \bf 53.4 \\
\bottomrule[1.28pt]
\end{tabular}
}
\label{tab:panoptic}
\end{table}

\subsection{Object Detection}

Object detection concurrently tackles the `where-and-what' problem, involving the localization of objects through bounding boxes and the assignment of class labels. This dual objective necessitates both accurate spatial localization and precise object classification. 
Given the prevalence of FPN-like architectures in many existing object detectors, the role of feature fusion becomes crucial for obtaining semantically consistent and clear feature maps, thereby enhancing the overall performance of the model in terms of both localization and classification.

\subsubsection{Experimental Settings}
\vspace{+1mm}
\noindent{\bf Datasets and Metrics.}
For object detection experiments, we leverage the MS COCO~\cite{2014microsoft} dataset, encompassing 80 object categories. Evaluation is performed using the Average Precision (AP) metric. The standard COCO metrics, including $AP$ (averaged over IoU thresholds from 0.5 to 0.95 with a stride of 0.05), $AP_{50}$ (IoU threshold is 0.5), $AP_{75}$ (IoU threshold is 0.75), $AP_{S}$, $AP_{M}$, and $AP_{L}$ are employed. Here, $S$, $M$, and $L$ denote small (area: 10-144 pixels), medium (area: 144 to 1024 pixels), and large objects (area: 1024 pixels and above), respectively.

\vspace{+1mm}
\noindent{\bf Implementation Details.}
Among the various existing detectors, we opt for the widely-used Faster R-CNN~\cite{fasterRCNN2015} with ResNet-50 and ResNet-101~\cite{resnet2016} as our baseline. Over the years, Faster R-CNN has undergone multiple design iterations, demonstrating stable performance and significant improvements since its original version. We choose to validate FreqFusion based on the Faster R-CNN architecture. The implementation provided by mmdetection~\cite{mmdetection} is employed, following its 1$\times$ (12 epochs) training configurations. Modifications are exclusively made to the feature fusion stages in the Feature Pyramid Network (FPN).

\subsubsection{Object Detection Results}
Quantitative and qualitative results are shown in Table~\ref{tab:FasterRCNN} and Figure~\ref{fig:coco}.
FreqFusion demonstrates the highest performance, achieving a notable 1.9 AP improvement on the COCO dataset, surpassing all competing methods, including CARAFE, IndexNet, A2U, FADE, SAPA-B, Dysample-S+, and Dysample+.

Notably, FreqFusion exhibits a lead over the second-place Dysample+, achieving a substantial margin of 0.7 AP. Even with ResNet-50 as the backbone, FreqFusion maintains competitive performance compared to the more robust ResNet-101, achieving comparable results at 39.4 AP.

When ResNet-101 is utilized as the backbone, FreqFusion continues to deliver a commendable 1.6 AP improvement, outperforming Dysample+ by 0.5 AP. These results underscore the robustness and efficacy of FreqFusion in enhancing object detection performance.

\subsection{Instance Segmentation}
Instance segmentation is a task involving the detection and delineation of each distinct object within an image. 
The intricate nature of instance segmentation demands not only the preservation of consistent category information but also the precise delineation of object boundaries. Therefore, the choice and quality of feature fusion become critical components in the design and evaluation of instance segmentation models.

\subsubsection{Experimental Settings}
\vspace{+1mm}
\noindent{\bf Datasets and metric.}
Similar to object detection, we leverage the widely used MS COCO~\cite{2014microsoft} dataset for our instance segmentation experiments. The evaluation metrics include the standard Box AP (Average Precision) and Mask AP, offering a comprehensive assessment of detection and segmentation performance.

\vspace{+1mm}
\noindent{\bf Implementation details}
For our instance segmentation experiments, we employ Mask RCNN~\cite{MaskRCNN2017} with both ResNet-50 and ResNet-101~\cite{resnet2016} as the baseline architectures. Following a similar approach to Faster R-CNN, our modifications are confined to the feature fusion stages within the Feature Pyramid Network (FPN). The codebase from mmdetection~\cite{mmdetection} is utilized, and training adheres to the default 1$\times$ schedule, comprising 12 epochs.
For Faster R-CNN with FPN, which fuse 4$\times$, 8$\times$, 16$\times$, and 32$\times$ downsampled features, we utilize 3 FreqFusion modules.

\subsubsection{Instance Segmentation Results}
Quantitative and qualitative results are shown in Table~\ref{tab:MaskRCNN} and Figure~\ref{fig:coco}, with ResNet serving as the backbone. 
FreqFusion showcases exceptional performance, manifesting a noteworthy improvement of 1.3 mask AP and 1.7 box AP on the COCO dataset. This accomplishment positions FreqFusion as the leading method, outperforming prominent competitors, including CARAFE, IndexNet, A2U, FADE, SAPA-B, Dysample-S+, and Dysample+.

FreqFusion demonstrates a substantial lead over the second-place Dysample+, achieving a margin of 0.3 mask AP and 0.4 box AP. Even when employing ResNet-50 as the backbone, FreqFusion maintains competitive performance compared to the more robust ResNet-101, yielding comparable results at 36.0 mask AP and 40.0 box AP.
With ResNet-101 as the backbone, FreqFusion consistently delivers commendable improvements, achieving a 1.4 mask AP and 1.6 box AP boost. This outperformance is highlighted by a 0.6/0.6 mask/box AP lead over Dysample+. These results underscore the robustness and efficacy of FreqFusion in advancing instance segmentation performance.

\subsection{Panoptic Segmentation}
Panoptic segmentation serves as a comprehensive integration of semantic segmentation and instance segmentation, providing a holistic perspective for classifying both stuff and things at the pixel level. In this section, we investigate the impact of various feature fusion methods on the panoptic segmentation task.

\subsubsection{Experimental Settings}
\vspace{+1mm}
\noindent{\bf Datasets and metrics.}
For panoptic segmentation, we utilize the MS COCO~\cite{2014microsoft} dataset, which encompasses 80 object categories. In this context, we report task-specific metrics, namely PQ, SQ, and RQ~\cite{2019panoptic}, as our evaluation criteria.

\vspace{+1mm}
\noindent{\bf Implementation details.}
From the panoptic segmentation model, we opt for Panoptic FPN~\cite{2019panopticfpn} with ResNet-50, modifying solely the upsampling stages in FPN. The mmdetection~\cite{mmdetection} codebase is employed, and we adhere to 1$\times$ (12 epochs) training configurations.
For Panoptic FPN, which fuse 4$\times$, 8$\times$, 16$\times$, and 32$\times$ downsampled features, we utilize 3 FreqFusion modules.

{\color{revise}
\subsubsection{Panoptic Segmentation Results}
To establish a clear and controlled baseline, we maintained the same settings as the previous work~\cite{2023dysample}. 
This allows us to fairly evaluate the performance improvements attributed to FreqFusion and compare it with recent state-of-the-art feature fusion methods. 

Table~\ref{tab:panoptic} provides a comprehensive overview of the quantitative results, with ResNet serving as the backbone. 
Qualitative results are shown in Figure~\ref{fig:coco}.
FreqFusion demonstrates exceptional performance, showcasing a significant improvement of 2.5 PQ on the COCO dataset. This places FreqFusion as the leading method, surpassing notable competitors, including CARAFE, IndexNet, A2U, FADE, SAPA-B, Dysample-S+, and Dysample+.

Notably, FreqFusion establishes a substantial lead over the second-place Dysample+, achieving a remarkable margin of 1.2 PQ. Even when employing ResNet-50 as the backbone, FreqFusion maintains competitive performance compared to the more robust ResNet-101, achieving higher results at 42.7 PQ as opposed to 42.2 PQ.

With ResNet-101 as the backbone, FreqFusion consistently delivers commendable improvements, achieving a notable 1.8 PQ boost. A 1.0 AP lead over Dysample+ further emphasizes this outperformance. These results strongly underscore the robustness and efficacy of FreqFusion in advancing Panoptic Segmentation performance.
}

\subsection{Ablation Studies}
In this section, we conduct ablation studies for the proposed FreqFusion.
We use the recent state-of-the-art efficient segmentation architecture SegNeXt~\cite{segnext} as the baseline for its high performance and efficiency.
Results are reported on challenging dataset ADE20K~\cite{ade20k}.

\noindent{\bf Ablation study for three generators.}
Table~\ref{tab:final_fusion} presents a comprehensive ablation study, meticulously assessing the individual contributions of the Adaptive Low-Pass Filter (ALPF), Adaptive High-Pass Filter (AHPF), and offset generators in the final fusion process of FreqFusion.
When ALPF is introduced independently, a marginal yet noteworthy improvement in mIoU of +0.9 is observed. This indicates the initial impact of adaptive low-pass filtering on refining high-level features during upsampling.
The combination of ALPF and AHPF demonstrates a synergistic effect, resulting in a more substantial enhancement. The mIoU achieves 42.9 (+1.8), highlighting the complementary roles played by both generators in addressing intra-category inconsistency and boundary displacement.
Notably, the inclusion of the offset generator leads to the highest mIoU of 43.5 (+2.4). 
These findings underscore the collaborative effectiveness of ALPF, AHPF, and offset generators within FreqFusion. Each component is instrumental in tackling specific challenges related to pixel-wise category consistency and spatial boundary accuracy.
Quantitative analyses in Table~\ref{tab:sim_analysis} also verify that three generators benefit solving intra-category consistency and boundary displacement.

\vspace{+1mm}
\noindent{\bf Ablation study for kernel sizes of adaptive filters.}
As shown in Table~\ref{tab:kernel_size}, we also investigate the kernel sizes of adaptive low/high-pass filters, \ie, $\bar K$ and $\hat K$.
When $\bar K=3$ and $\hat K=3$, the proposed FreqFusion improves upon SegNeXt~\cite{segnext} by 0.8 mIoU.
Increasing the kernel size of adaptive low-pass filters from $\bar K=3$ to $\hat K=5$ results in an additional 1.0 mIoU improvement, and performance remains consistent from $\bar K=5$ to $\bar K=7$.
However, further increasing the kernel size of adaptive high-pass filters from $\hat K=3$ to $\hat K=5$ leads to performance degradation, \ie, from 42.9 to 42.4.
Thus, we set $\bar K=5$ and $\hat K=3$.

\vspace{+1mm}
\noindent{\bf Ablation study for offset generator.}
Here, we further investigate the setting of offset groups. Features are divided into different groups along the channel dimension, and different groups of offsets are generated, sharing the same sampling set in each group to achieve finer resampling. 
As presented in Table~\ref{tab:offset_group}, the results demonstrate that utilizing four offset groups achieves the highest mIoU of 43.5, indicating that this configuration yields the most effective refinement of segmentation predictions. Increasing the number of offset groups beyond four does not lead to significant improvements and may even result in marginal performance degradation.

\vspace{+1mm}
\noindent{\bf Ablation study for initial fusion.}
Despite numerous works acknowledging the problem of blurred boundaries caused by simple interpolation methods, many still resort to using such techniques as bilinear interpolation for initial fusion~\cite{fade, 2023sfnet}.
The results in Table~\ref{tab:initial_fusion} demonstrate the effectiveness of the enhanced initial fusion in improving segmentation performance. When solely utilizing the ALPF generator, there is a noticeable boost of +0.3 in mIoU. Furthermore, the combination of both ALPF and AHPF generators in the enhanced initial fusion further elevates the performance to 43.5. This highlights the collaborative synergy of both generators in refining segmentation features during the initial fusion process.
The findings underscore the significance of the initial fusion stage for achieving superior segmentation results. By addressing the inherent limitations of simple interpolation methods, such as blurred boundaries, we can significantly enhance the accuracy and quality of segmentation outcomes.

\section{Discussion with Related Works}
\label{sec:discussion}
Our work is closely related to previous studies in various aspects, and we elaborate on these relationships and distinctions in detail.

Kernel-based methods like A2U~\cite{2021a2u} and IndexNet~\cite{2020indexnet} exclusively rely on low-level features for dynamic kernel generation, posing a risk of introducing noise into the kernels. Similarly, SAPA~\cite{sapa} leverages the similarity between low-level and high-level features for kernel generation, but this approach also carries the potential for noise introduction.
In contrast, CARAFE~\cite{carafe} solely employs high-level features for dynamic kernel generation, overlooking the importance of the high-resolution structure inherent in low-level features, which has been demonstrated to be crucial for effective upsampling~\cite{fade}. Conversely, FADE~\cite{fade} incorporates both low-level and high-level features for dynamic kernel generation. However, it still adopts simple nearest-neighbor interpolation for upsampling, leading to boundary displacement issues.
To address these challenges, our proposed FreqFusion utilizes both low-level and high-level features and introduces the ALPF generator and AHPF generator to enhance the initial fusion.

Recent sampling-based methods such as AlignSeg~\cite{2021alignseg}, IFA~\cite{2022IFA}, SFNet~\cite{2023sfnet}, FaPN~\cite{2021fapn}, and Dysample~\cite{2023dysample} predominantly concentrate on enhancing upsampling by learning to sample features with potential feature inconsistency.
In contrast, FreqFusion adopts a novel approach. It initially smooths the high-level features to reduce overall feature inconsistency, then utilizes local similarity as guidance, and finally learns to resample the features to replace inconsistent features.

While existing kernel-based and sampling-based methods primarily focus on improving the upsampling process in feature fusion, FreqFusion goes a step further by extracting high-frequency information from low-level features and incorporating it residually to enhance feature fusion. 
Furthermore, although previous studies empirically observe the problems in standard feature fusion and attempt to address them, they lack clear definitions supported by quantitative measurements. In contrast, we precisely identify and define the issues of intra-category inconsistency and boundary displacement, measuring them through feature similarity analysis. The proposed FreqFusion effectively tackles these issues with the aim of achieving simultaneous feature consistency and boundary sharpness.

\begin{table}[tb!]
\centering
\caption{
Ablation study for the final fusion of FreqFusion.
Results are reported on ADE20K~\cite{ade20k} validation set.
}
\scalebox{0.6918}{
\begin{tabular}{c|ccc|c|c|cc}
\toprule[1.28pt]
\multirow{2}{*}{SegNeXt-T~\cite{segnext}} &ALPF & AHPF & Offset & Params &FLOPs  & \multirow{2}{*}{mIoU} \\
    & generator  & generator & generator & (M) & (G)\\
\midrule
Standard feature fusion  & $\times$ & $\times$ & $\times$ &4.26 &6.59 &41.1 \\
\midrule
\multirow{3}{*}{FreqFusion} & $\surd$ & $\times$ & $\times$ & 4.42 & 6.91 & 42.0 (+0.9)\\ 
& $\surd$ & $\surd$ & $\times$ &4.43 &6.97 & 42.9 (+1.8) \\
& $\surd$ & $\surd$ & $\surd$ &4.59 & 7.14 &\bf 43.5 (+2.4)\\
\bottomrule[1.28pt]
\end{tabular}
}
\label{tab:final_fusion}
\end{table}

\begin{table}[tb!]
\centering
\caption{
Ablation study for the FreqFusion.
Results are reported on ADE20K~\cite{ade20k} validation set.
$\bar K$, $\hat K$ indicates the kernel size of adaptive low/high-pass filters.
}
\scalebox{0.8918}{
\begin{tabular}{c|cc|cc|cc}
\toprule[1.28pt]
\multirow{2}{*}{SegNeXt-T~\cite{segnext}} &ALPF & AHPF & \multirow{2}{*}{mIoU} \\
 & generator  & generator \\
\midrule
Standard feature fusion & $\times$ & $\times$ &41.1 \\
\midrule
\multirow{4}{*}{FreqFusion} & $\bar K=3$ & $\hat K=3$ & 
 41.9 (+0.8) \\
& $\bar K=5$ & $\hat K=3$  &\bf 42.9 (+1.8) \\
& $\bar K=7$ & $\hat K=3$  &\bf 42.9 (+1.8) \\ 
& $\bar K=5$ & $\hat K=5$  &  42.4 (+1.3) \\ 
\bottomrule[1.28pt]
\end{tabular}
}
\label{tab:kernel_size}
\end{table}

\begin{table}[tb!]
\centering
\caption{
Ablation study on the number of offset groups for the offset generator.
Results are reported on ADE20K~\cite{ade20k} validation set.
}
\scalebox{0.928}{
\begin{tabular}{c|c|c|c|c|ccc}
\toprule[1.28pt]
Offset group & 1 & 2 & 4 & 8 & 16 \\
\midrule
mIoU & 43.1 & 43.2 &\bf 43.5& 43.4 & 43.3\\ 
\bottomrule[1.28pt]
\end{tabular}
}
\label{tab:offset_group}
\end{table}

\begin{table}[tb!]
\centering
\caption{
Ablation study for the initial fusion of FreqFusion.
Results are reported on ADE20K~\cite{ade20k} validation set.
The generators in the initial fusion share the same parameters as those in the final fusion, thus the enhanced initial fusion introduces no extra parameters.
}
\scalebox{0.918}{
\begin{tabular}{c|cc|cc|ccc}
\toprule[1.28pt]
\multirow{2}{*}{SegNeXt-T~\cite{segnext}} &ALPF & AHPF & \multirow{2}{*}{mIoU} \\
    & generator  & generator \\
\midrule
FreqFusion  & $\times$ & $\times$ &43.1 \\
\midrule
\multirow{1}{*}{FreqFusion} & $\surd$ & $\times$ & 43.4 \\ 
\multirow{1}{*}{w/ Enhanced initial fusion}& $\surd$ & $\surd$ & 43.5 \\
\bottomrule[1.28pt]
\end{tabular}
}
\label{tab:initial_fusion}
\vspace{-2.918mm}
\end{table}

\section{Conclusion}
\label{sec:conclusion}
{\color{revise}
In this paper, we aim to address the critical challenges of intra-category inconsistency and boundary displacement in dense image prediction tasks.
By employing feature similarity analysis, we quantitatively measured these issues, guiding the design of the proposed method, Frequency-aware Feature Fusion (FreqFusion). 
FreqFusion comprises an Adaptive Low-Pass Filter (ALPF) generator, an Offset generator, and an Adaptive High-Pass Filter (AHPF) generator. These components effectively tackle intra-category inconsistency and boundary displacement by adaptively smoothing high-level features, resampling nearby category-consistent features, and enhancing the high frequencies of lower-level features. Through qualitative and quantitative evaluations, we have demonstrated the superiority of FreqFusion across various dense prediction tasks, including semantic segmentation, object detection, instance segmentation, and panoptic segmentation.

In the future, investigating the computational efficiency and real-time applicability of FreqFusion in resource-constrained environments will be essential for its practical deployment. Moreover, extending FreqFusion to incorporate temporal inputs, such as videos, where temporal intra-category consistency and boundary sharpness may be disrupted by motion blur or occlusion, could further enhance its adaptability to diverse visual perception tasks like object tracking.
}

\section*{Acknowledgments}
This work was supported by the National Natural Science Foundation of China (62331006, 62171038, 61936011, and 62088101), the R\&D Program of Beijing Municipal Education Commission (KZ202211417048), the Fundamental Research Funds for the Central Universities, and JST Moonshot R\&D Grant Number JPMJMS2011, Japan.

{
\small
\bibliographystyle{IEEEtran}
\bibliography{./egbib.bib}
}

\begin{IEEEbiography}[{\includegraphics[width=1in,height=1.25in,clip,keepaspectratio]{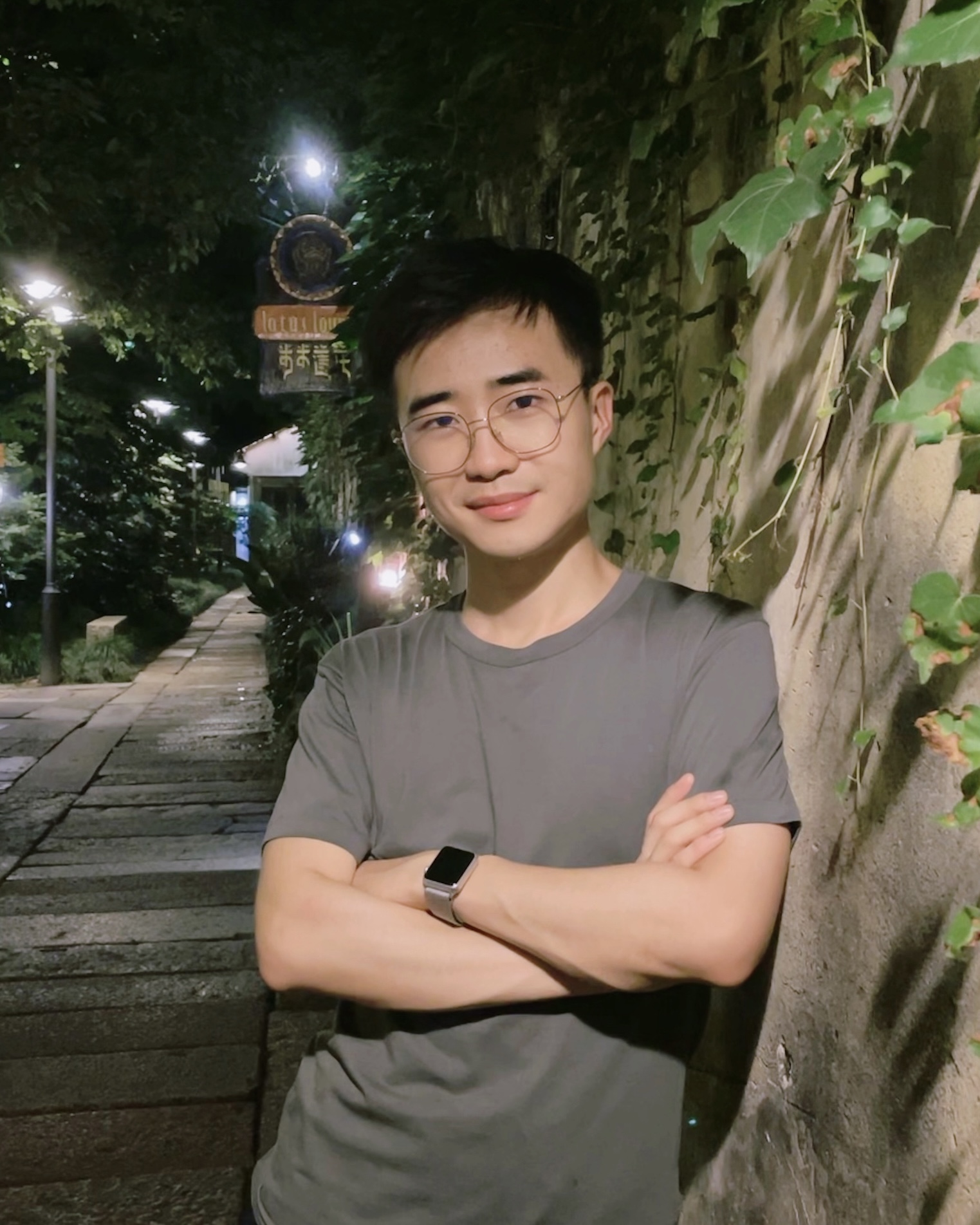}}]{Linwei Chen}
received the B.S. degree in mechanical engineering and automation from the China University of Geosciences, Beijing, China, in 2019, and the M.S. degree in software engineering from the Beijing Institute of Technology, Beijing, China, in 2021.
He is currently Eng.D. at MIIT Key Laboratory of Complex-field Intelligent Sensing, the School of Information and Electronics, Beijing Institute of Technology.
His research interests include image segmentation, object detection, and remote sensing.
\end{IEEEbiography}

\begin{IEEEbiography}[{\includegraphics[width=1in,height=1.25in,clip,keepaspectratio]{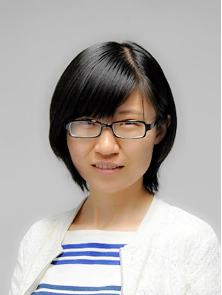}}]{Ying Fu}
received the B.S. degree in Electronic Engineering from Xidian University in 2009, the M.S. degree in Automation from Tsinghua University in 2012, and the Ph.D. degree in information science and technology from the University of Tokyo in 2015. 
She is currently a professor at the School of Computer Science and Technology, Beijing Institute of Technology. Her research interests include physics-based vision, image and video processing, and computational photography. 
\end{IEEEbiography}

\begin{IEEEbiography}[{\includegraphics[width=1in,height=1.25in,clip,keepaspectratio]{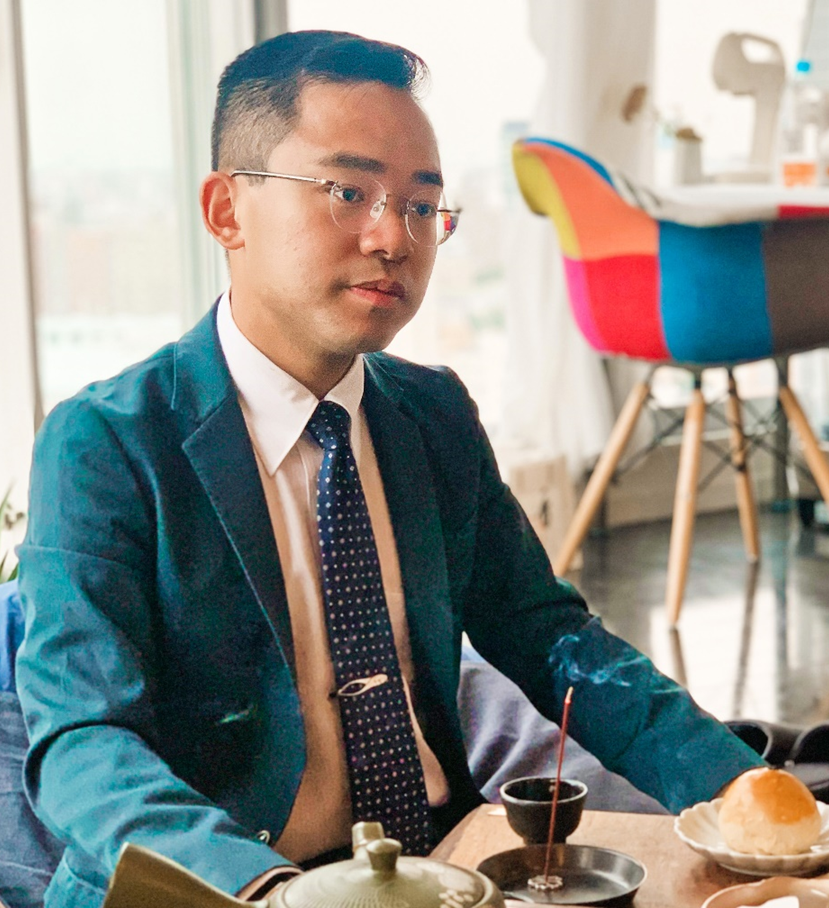}}]{Lin Gu}
completed his Ph.D. studies at the Australian National University and NICTA (Now Data61) in 2014.  After that, he was associated with the National Institute of Informatics in Tokyo and the Bioinformatics Institute, A*STAR, Singapore. Currently, he is now a research scientist at RIKEN AIP, Japan, and a special researcher at the University of Tokyo. He is also a project manager for Moonshot R\&D and the RIKEN-MOST program. His research covers a wide range of topics, encompassing computer vision, medical imaging, and AI for science.
\end{IEEEbiography}

\begin{IEEEbiography}[{\includegraphics[width=1in,height=1.25in,clip,keepaspectratio]{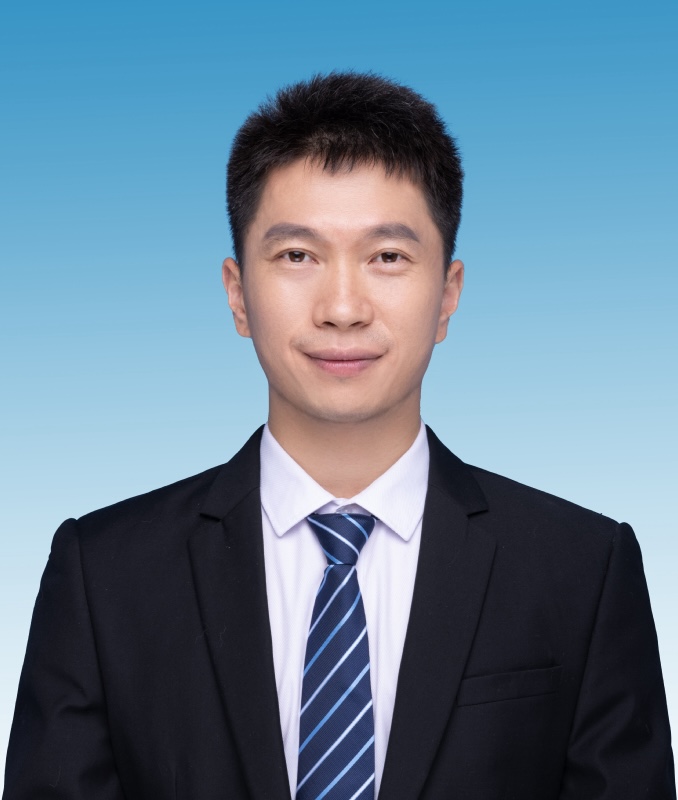}}]{Chenggang Yan}
received the B.S. degree in Control Science and Engineering from Shandong University, Shandong, China, in 2008, and the Ph.D. degree in Computer Science from the Chinese Academy of Sciences University, Beijing, China, in 2013. He is currently a professor in the Department of Automation at Hangzhou Dianzi University. His research interests include computational photography, pattern recognition, and intelligent systems.
\end{IEEEbiography}

\begin{IEEEbiography}[{\includegraphics[width=1in,height=1.25in,clip,keepaspectratio]{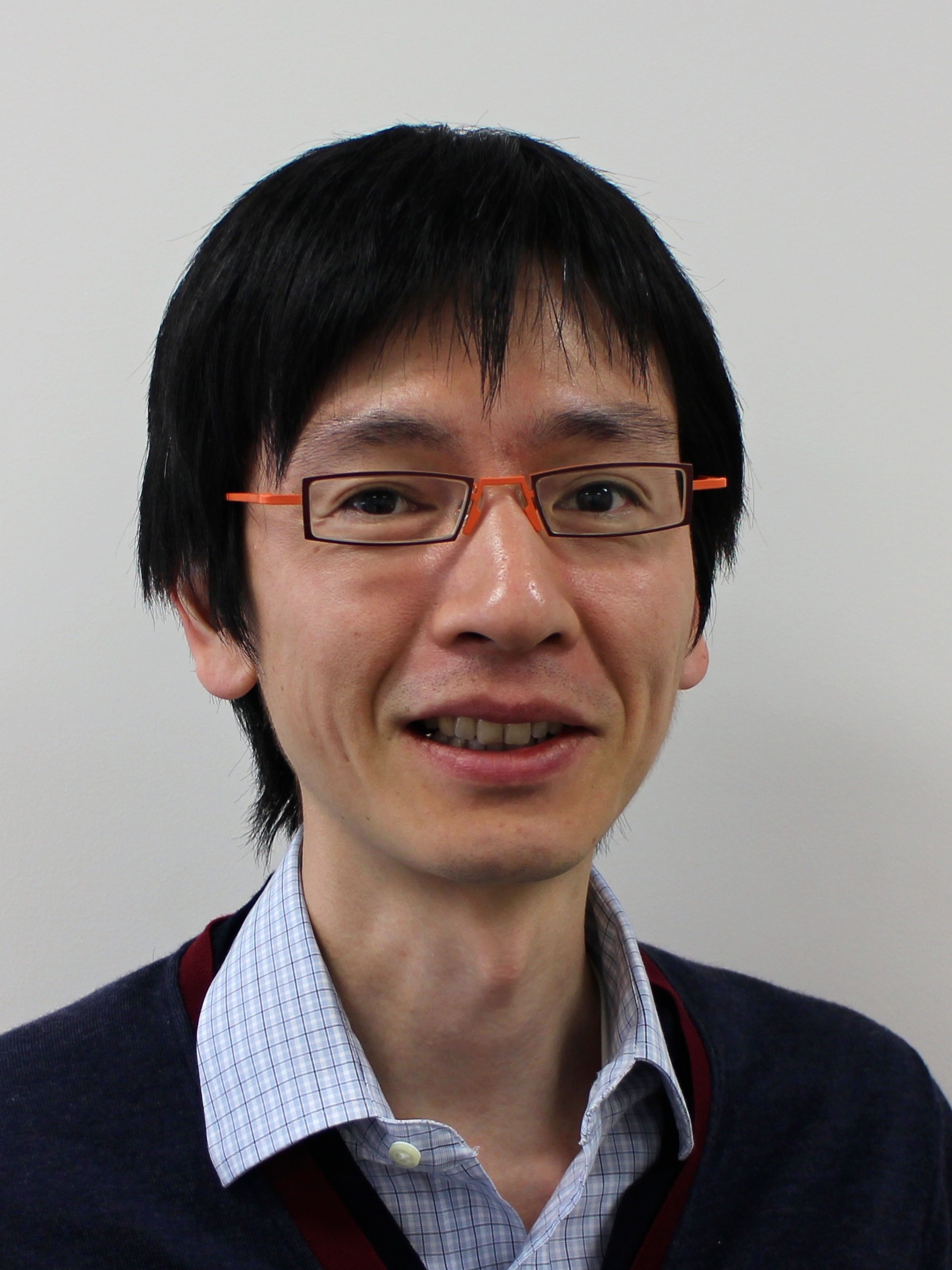}}]{Tatsuya Harada}
received the PhD degree in mechanical engineering from the University of Tokyo, in 2001. He is currently a professor with the Research Center for Advanced Science and Technology, the University of Tokyo, a team leader with the RIKEN Center for Advanced Intelligence Project (AIP), and a visiting professor with the National Institute of Informatics (NII). His research interests include visual recognition, machine learning and intelligent robot
\end{IEEEbiography}

\begin{IEEEbiography}[{\includegraphics[width=1in,height=1.25in,clip,keepaspectratio]{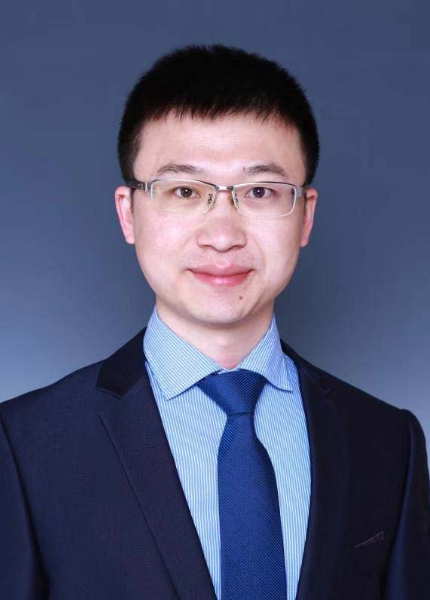}}]{Gao Huang}
is an Associate Professor in the Department of Automation at Tsinghua University. Previously, he served as a Postdoctoral Researcher in the Department of Computer Science at Cornell University. He obtained his PhD degree in Control Science and Engineering from Tsinghua University in 2015, and his B.Eng degree in Automation from Beihang University in 2009. He also undertook visiting student positions at Washington University in St. Louis and Nanyang Technological University in 2013 and 2014, respectively. His research interests include machine learning and computer vision.
\end{IEEEbiography}

\end{document}